\documentclass{article}
\usepackage{ijcai26}
\usepackage{times}
\usepackage{soul}
\usepackage{url}
\usepackage[hidelinks]{hyperref}
\usepackage[utf8]{inputenc}
\usepackage[small]{caption}
\usepackage{subcaption}
\usepackage{pifont}
\usepackage{graphicx}
\usepackage{amsmath}
\usepackage{amssymb}
\usepackage{amsthm}
\usepackage{booktabs}
\usepackage{algorithm}
\usepackage{algorithmic}
\usepackage{pgfplots}
\usepackage{threeparttable}
\usepackage{multirow}
\usepackage[switch]{lineno}
\usepackage{enumitem}

\newif\ifanonymous
\anonymousfalse % Set to false for camera-ready / arxiv

\pgfplotsset{compat=1.18}
\definecolor{r50}{HTML}{1A237E}
\definecolor{r60}{HTML}{B71C1C}
\definecolor{r70}{HTML}{1B5E20}
\definecolor{r80}{HTML}{3E2723}

\newtheorem{definition}{Definition}

\title{DiSGMM: A Method for Time-varying Microscopic Weight Completion on\\Road Networks}

\ifanonymous
\linenumbers
\author{
First Author$^1$
\and
Second Author$^2$\and
Third Author$^{2,3}$\And
Fourth Author$^4$\\
\affiliations
$^1$First Affiliation\\
$^2$Second Affiliation\\
$^3$Third Affiliation\\
$^4$Fourth Affiliation\\
\emails
\{first, second\}@example.com,
third@other.example.com,
fourth@example.com
}
\else
\author{
Yan Lin$^1$
\and
Jilin Hu$^2$\and
Shengnan Guo$^3$\and
Christian S. Jensen$^1$
\and\\
Youfang Lin$^3$
\and
Huaiyu Wan$^3$\\
\affiliations
$^1$Aalborg University\\
$^2$East China Normal University\\
$^3$Beijing Jiaotong University\\
\emails
\{lyan, csj\}@cs.aau.dk,
jlhu@dase.ecnu.edu.cn,
\{guoshn, yflin, hywan\}@bjtu.edu.cn
}
\fi

\begin{document}

\maketitle

\begin{abstract}
Microscopic road-network weights represent fine-grained, time-varying traffic conditions obtained from individual vehicles. An example is travel speeds associated with road segments as vehicles traverse them. These weights support tasks including traffic microsimulation and vehicle routing with reliability guarantees.
We study the problem of time-varying microscopic weight completion. During a time slot, the available weights typically cover only some road segments. Weight completion recovers distributions for the weights of every road segment at the current time slot.
This problem involves two challenges: (i) contending with two layers of sparsity, where weights are missing at both the network layer (many road segments lack weights) and the segment layer (a segment may have insufficient weights to enable accurate distribution estimation); and (ii) achieving a weight distribution representation that is closed-form and can capture complex conditions flexibly, including heavy tails and multiple clusters.

To address these challenges, we propose DiSGMM that combines sparsity-aware embeddings with spatiotemporal modeling to leverage sparse known weights alongside learned segment properties and long-range correlations for distribution estimation. DiSGMM represents distributions of microscopic weights as learnable Gaussian mixture models, providing closed-form distributions capable of capturing complex conditions flexibly. Experiments on two real-world datasets show that DiSGMM can outperform state-of-the-art methods.
\end{abstract}

\section{Introduction}
\begin{figure}[t]
    \centering
    \begin{subfigure}{0.45\linewidth}
        \centering
        \includegraphics[width=\linewidth]{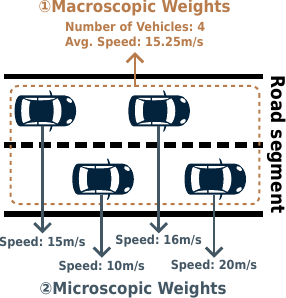}
        \caption{Macroscopic and microscopic weights.}
        \label{fig:two-types-weights}
    \end{subfigure}
    \hfill
    \begin{subfigure}{0.52\linewidth}
        \centering
        \includegraphics[width=\linewidth]{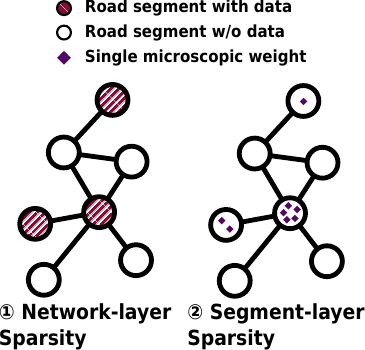}
        \caption{Two layers of data sparsity.}
        \label{fig:two-layers-sparsity}
    \end{subfigure}
    \caption{Motivation illustration.}
    \label{fig:motivation}
\end{figure}

Road networks are essential to modern transportation~\cite{DBLP:journals/tkde/GaoQZ015,DBLP:conf/kdd/Fang0ZHCGJ22}.
A road network is usually represented as a directed graph $\mathcal G=(\mathcal V, \mathcal E)$, where nodes $v_i \in \mathcal V$ model road intersections or endpoints and edges $e_i \in \mathcal E$ model road segments. The edges are associated with weights that describe traffic conditions on the road network, which we term road network weights.
Two types of such weights exist: \textit{macroscopic weights} and \textit{microscopic weights}, as exemplified in Figure~\ref{fig:two-types-weights}.
Macroscopic weights describe aggregated traffic conditions, including numbers of vehicles on a road segment or the average speed of vehicles.
Microscopic weights capture the vehicle-level traffic conditions, including individual vehicle's speed when traveling along a road segment or its travel time when traversing a road segment.

Macroscopic weights capture general traffic conditions on a road network and support tasks such as traffic flow~\cite{DBLP:conf/aaai/GuoLFSW19} and congestion~\cite{akhtar2021review} prediction.
While existing studies~\cite{tedjopurnomo2020survey,jiang2022graph,jin2023spatio} consider macroscopic weights, we target microscopic weights that capture fine-grained, vehicle-level traffic conditions and support tasks such as traffic microsimulation~\cite{hollander2008principles}, driving behavior analysis~\cite{abou2020application}, and routing with reliability guarantees~\cite{DBLP:conf/icde/HuG0J19}.
More specifically, we focus on the problem of \textit{time-varying microscopic weight completion}.
The input is a road network and sparse microscopic weights observed within the current and historical time slots, where many segments lack weights within a time slot and segments with weights may have only few of them.
The goal is to recover the accurate representation of weights for every road segment within the current time slot.
This problem presents two unique challenges.

\textbf{Solutions must contend with two layers of data sparsity.}
First, we have \textit{network-layer sparsity}, as shown in Figure~\ref{fig:two-layers-sparsity}\ding{172}. In any time slot, the available vehicle movement data can be limited and thus may cover only a fraction of the entire road network. In other words, some road segments will have no microscopic weights during a time slot.
Contending with this sparsity will require combining multiple information sources, including long-range spatial neighbors and long-term temporal history that include weights, as well as inherent information of road segments such as their typical microscopic weight distribution, in order to recover weights for road segments without weights.

Second, we have \textit{segment-layer sparsity}, as shown in Figure~\ref{fig:two-layers-sparsity}\ding{173}. A road segment with microscopic weights may still contain too few weights to enable estimation of an accurate representation of the microscopic weights of the segment.
This calls for sparsity-awareness to decide whether to rely more on available weights in case of low sparsity or to rely more on other information sources in case of high sparsity.
Existing methods~\cite{DBLP:conf/icde/HuG0J19,DBLP:conf/gis/Muniz-CuzaHZP022,DBLP:conf/sdm/HanCGMM022,yuan2024nuhuo} either disregard segment-layer sparsity or simply discard all available data in segments with high sparsity, which increases network-layer sparsity.

\begin{figure}[t]
    \centering
    \includegraphics[width=\linewidth]{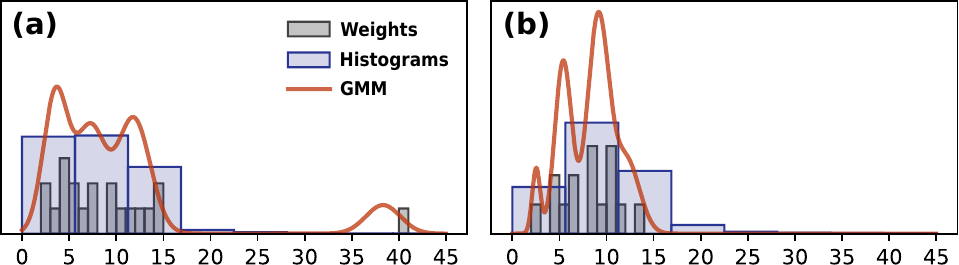}
\caption{Two cases of representing microscopic weights with 8-bin histograms versus 4-component Gaussian mixture models (GMM).}
    \label{fig:hist-vs-gmm}
\end{figure}

\textbf{Solutions must represent microscopic weights judiciously.}
As discussed above, microscopic weights capture vehicle-level traffic conditions. However, representing them directly as the set of all individual weights is inefficient, since the number of observed vehicles varies considerably across time and segments, making it difficult to derive accurate and generalizable representations from sparse weights.
Rather, we need a representation that can capture complex traffic conditions accurately, including distributions with heavy tails and multiple clusters, while remaining interpretable and efficient.

Most existing methods~\cite{DBLP:conf/icde/HuG0J19,DBLP:conf/gis/Muniz-CuzaHZP022,DBLP:conf/sdm/HanCGMM022,yuan2024nuhuo} adopt histograms with fixed range and number of bins to represent microscopic weight distributions. However, such histograms face a trade-off between range coverage and resolution.
For example, as shown in Figure~\ref{fig:hist-vs-gmm}, consider sets of microscopic weights with multiple clusters in the range $[0, 15]$. In case (a), a heavy tail value $41$ forces the histogram to cover the range $[0, 41]$, reducing resolution where most weights lie. In both cases, the fixed range and number of bins makes it difficult for histograms to distinguish the multiple clusters accurately, while Gaussian mixture models can capture these patterns flexibly.
Diffusion models~\cite{ho2020denoising,wen2023diffstg} are alternatives that can flexibly model complex distributions. However, they do not provide closed-form representations, making them impractical for microscopic weight completion tasks that require explicit distribution information.

We propose \textit{\underline{Di}-\underline{S}parsity \underline{G}raph Gaussian \underline{M}ixture \underline{M}odel} (\textbf{DiSGMM}) for accurate time-varying microscopic weight completion.
Design choices in DiSGMM address the above two challenges.
First, to contend with the two layers of data sparsity, DiSGMM introduces two modules for learning latent embeddings of different information sources: one module for learning a static embedding for each road segment that incorporates inherent information; one module for learning a dynamic embedding for each road segment at a certain time slot, balancing known weights and inherent information with awareness of the segment-layer sparsity.
DiSGMM also incorporates a spatiotemporal U-net with global residual connections, which is effective at gathering information from long-range spatial neighbors and long-term temporal history, thus contending with network-layer sparsity.
Second, for the representation of microscopic weights, DiSGMM adopts a learnable Gaussian mixture model~\cite{reynolds2009gaussian}, which provides closed-form distributions while being flexible enough to capture complex traffic conditions including ones with heavy tails and multiple clusters.
We report on extensive experiments on two real-world datasets that compare the performance of DiSGMM against several state-of-the-art methods. Results providing evidence that DiSGMM achieves its design goal.

The paper's main contributions are summarized as follows:
\begin{itemize}[leftmargin=*]
  \item We propose DiSGMM, a novel method for time-varying microscopic weight completion on road networks that addresses two unique challenges of the problem.
  \item The method features a dual sparsity-aware architecture that combines static and dynamic embeddings with a spatiotemporal U-net, effectively handling both segment-layer and network-layer sparsity.
  \item The method enables learnable Gaussian mixture models for representing microscopic weights, providing closed-form distributions that flexibly represent complex traffic conditions with heavy tails and multiple clusters.
  \item We report on extensive experiments on two real-world datasets, finding that DiSGMM is able to consistently outperform state-of-the-art methods.
\end{itemize}

\section{Related Work}
Road network weight completion is a fundamental problem in traffic data mining, aiming to recover missing or incomplete weight information across road segments due to limitations in data collection.
As mentioned in the introduction, we categorize road network weights into macroscopic and microscopic weights.
Below, we briefly discuss macroscopic weight completion to provide context.
Then we discuss existing microscopic weight completion studies, which represent alternatives to our proposal.

\subsection{Macroscopic Weight Completion}

Macroscopic weight completion has been studied extensively due to its importance in traffic monitoring and prediction.
Existing methods can be categorized by their spatial-temporal modeling techniques.
Tensor completion methods~\cite{chen2021low,10.1145/3278607} leverage low-rank assumptions to capture global patterns in traffic data, such as daily and weekly periodicity.
RNN-based models~\cite{hochreiter1997long,wei2018brits,liu2019naomi} capture temporal dependencies among consecutive time steps through sequential processing.
GNN-based approaches~\cite{kipf2016semi,cini2021filling,wu2021inductive,DBLP:journals/kbs/WeiLGLZJWW24} explicitly model spatial correlations among road segments using road network structures.
Attention-based methods~\cite{shukla2021multi,nie2024imputeformer,DBLP:conf/nips/VaswaniSPUJGKP17} provide flexible long-range dependency modeling across both space and time.
GAN-based models~\cite{goodfellow2014generative,yoon2018gain,DBLP:conf/iclr/LiJM19} and diffusion models~\cite{tashiro2021csdi,liu2023pristi} offer probabilistic imputation with uncertainty quantification.

However, macroscopic and microscopic weight completion have fundamentally different objectives and data characteristics.
Macroscopic weights are aggregated measurements (e.g., average travel times, vehicle counts), and their completion focuses on recovering these aggregate statistics using available observations and spatiotemporal correlations.
In contrast, microscopic weights capture vehicle-level traffic conditions, and their completion requires recovering the full distribution of weights for each road segment and time slot, not just aggregated statistics.
Furthermore, microscopic weight completion faces unique challenges including the two layers of data sparsity and the need for appropriate representations of distributions, which are not concerns in macroscopic weight completion.

\subsection{Microscopic Weight Completion}

Compared to macroscopic weight completion, microscopic weight completion has received much less attention.
GCWC~\cite{DBLP:conf/icde/HuG0J19} is the first study to explicitly address microscopic weight completion. Instead of completing weights as single average values, GCWC completes weights as distributions in the form of histograms and uses a GCN-based network to complete the weight distributions of all road segments during a target time slot. While it considers spatial correlations between road segments, it disregards temporal correlations across different time slots.
SSTGCN~\cite{DBLP:conf/gis/Muniz-CuzaHZP022} addresses this limitation by employing TCNs~\cite{lea2016temporal} and GCNs to capture both spatial and temporal correlations between road segments. It also introduces a mechanism to avoid propagating noisy information when dealing with high sparsity.
ConGC~\cite{DBLP:conf/sdm/HanCGMM022} further improves completion accuracy on sparse data by utilizing a graph containing contextual information of road segments, such as speed limits and numbers of lanes, to achieve higher accuracy when only few weights are known. However, the contextual information needed to build such graphs is not always available.
Nuhuo~\cite{yuan2024nuhuo} partitions the road network into regions and uses a two-level encoding scheme: a global encoder captures inter-region correlations, while a local encoder models fine-grained spatial and temporal patterns within each region.

However, existing methods do not tackle the two challenges identified in the introduction.
First, while they address network-layer sparsity through spatial and temporal modeling, they do not explicitly handle segment-layer sparsity, failing to adaptively adjust their reliance on sparse weights versus other information sources.
Second, all existing methods use histograms to represent microscopic weights, which face a trade-off between range coverage and resolution that makes it difficult to represent complex distributions such as ones including heavy tails and multiple clusters.

\section{Preliminaries}

% We present key concepts and then the problem definition.

\begin{definition}
[Road Network]
A road network is modeled as a directed graph $\mathcal G=(\mathcal V, \mathcal E)$, where $\mathcal V$ is a set of nodes $v_i$ that represent road intersections or endpoints of road segments and where $\mathcal E$ is a set of directed edges $e_i = (v_j, v_k)$ that represent road segments.
Each edge $e$ has an associated length $l_e$ that captures the length of the segment it represents.
\end{definition}

\begin{definition}
[Vehicle-level Travel Speed]
For a road segment $e$ with length $l_e$, the vehicle-level travel speed is defined as $s = l_e / \Delta t$, where $\Delta t$ is the time elapsed between a vehicle entering and leaving the segment.
\end{definition}

\begin{definition}
[Time-varying Microscopic Weights]
In a road network $\mathcal G$, during a time slot $T=[t^1,t^2)$, each road segment $e \in \mathcal E$ is associated with a set of microscopic weights $\mathbb{W}_e^T$, where each element is the vehicle-level travel speed of a vehicle that traverses segment $e$ during time slot $T$.
We adopt travel speed rather than travel time as the weight because speed is normalized by segment length, enabling direct comparison across segments. In contrast, travel times scale with segment length.
\end{definition}

\noindent \textbf{Problem definition.}
\textit{Time-varying microscopic weight completion.}
Given a road network $\mathcal G=(\mathcal V, \mathcal E)$ and a sequence of historical time slots $\langle T_1, T_2, \dots, T_{H-1} \rangle$ along with the current time slot $T_H$, where $T_i=[t_i^1, t_i^2)$, we have microscopic weights $\{\mathbb{W}_e^{T_1}, \mathbb{W}_e^{T_2}, \dots, \mathbb{W}_e^{T_H}\}$ for each road segment $e \in \mathcal E$.
Note that some sets $\mathbb{W}_e^{T_i}$ may be empty or sparse due to limited data coverage.
The goal is to estimate an accurate representation of the microscopic weight distribution for element in $\mathbb{W}_e^{T_H}$ across all road segments in $\mathcal E$.

\section{Methodology}

\begin{figure*}[t]
    \centering
    \includegraphics[width=1.0\linewidth]{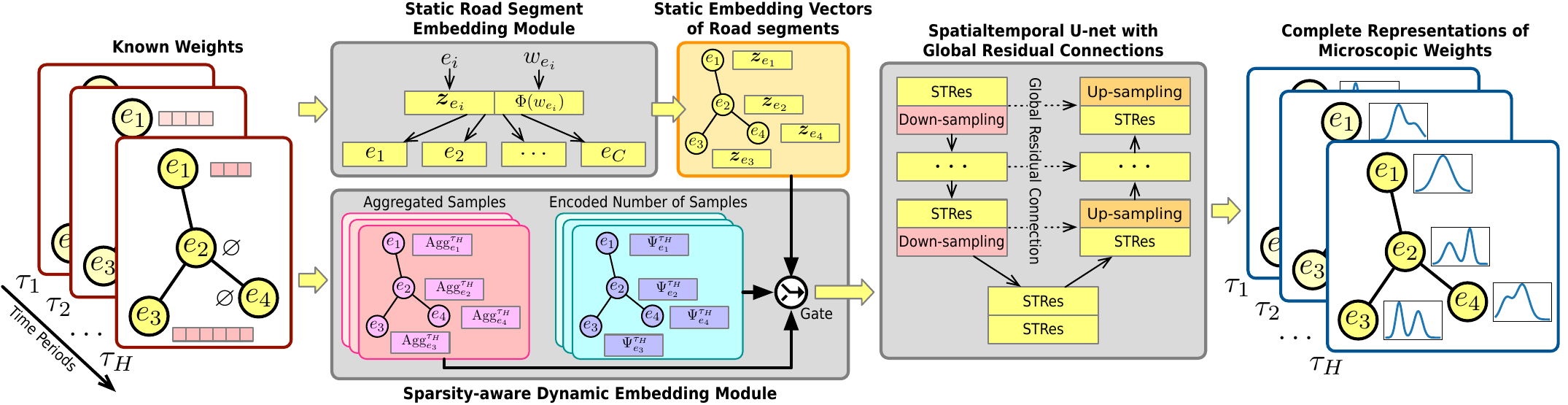}
    \caption{Overview of DiSGMM.}
    \label{fig:overall-framework}
\end{figure*}

As shown in Figure~\ref{fig:overall-framework}, DiSGMM addresses the two unique challenges of microscopic weight completion as follows.
For the first challenge of \textbf{two layers of sparsity}, DiSGMM incorporates three main modules: a \textit{static road segment embedding module} that captures inherent information of each road segment from network topology and historical weights; a \textit{sparsity-aware dynamic embedding module} that balances known weights and inherent information with awareness of segment-layer sparsity; and a \textit{spatiotemporal U-net with global residual connections} that gathers information from long-range spatial neighbors and long-term temporal history, contending with network-layer sparsity.
For the second challenge of \textbf{representation of microscopic weights}, DiSGMM adopts learnable Gaussian mixture models, providing closed-form distributions of microscopic weights that flexibly capture complex traffic conditions including heavy tails and multiple clusters.

The following sections detail each module of DiSGMM.

\subsection{Static Road Segment Embedding} \label{sec:embedding-module}

The microscopic weight distribution of a road segment is closely tied to its inherent properties. To capture such properties without relying on additional contextual information, we design a static embedding module that learns the static embedding vector of each road segment from two sources: (1) \textit{network topology}, reflecting a segment's role and neighboring structure, and (2) \textit{historical weight patterns}, since segments with similar historical microscopic weights typically share similar properties.

For topology, we perform random walks on the road network $\mathcal G$ following DeepWalk~\cite{DBLP:conf/kdd/PerozziAS14} to generate sequences of contextual segments around each target segment $e$. For historical patterns, we collect the microscopic weights of $e$ across all historical time slots.

We then employ an objective inspired by skip-gram~\cite{DBLP:journals/corr/abs-1301-3781} that maximizes the co-occurrence probability of contextual segments given the target segment and its historical weights. Specifically, we assign a static embedding vector $\boldsymbol z_e\in \mathbb R^d$ to each segment $e$, an output vector $\boldsymbol z'_{e_k} \in \mathbb R^{2d}$ to each contextual segment, and encode each historical weight $w_e$ using learnable Fourier features $\Phi(w_e) \in \mathbb R^d$~\cite{DBLP:conf/nips/LiSLHB21}. The objective is:
\begin{equation}
\mathcal L_{\mathrm{embed}} = -\log \prod_{c=1}^C \frac{\exp((\boldsymbol z_e || \Phi(w_e)) {\boldsymbol z'_{e_c}}^\top)}{\sum_{e_k\in \mathcal E} \exp((\boldsymbol z_e || \Phi(w_e))  {\boldsymbol z'_{e_k}}^\top)},
\label{eq:segment-embedding-loss}
\end{equation}
where $||$ denotes concatenation and $\{e_1, \ldots, e_C\}$ are contextual segments from the random walk. This ensures that segments with similar topology and historical weight distributions learn similar embeddings $\boldsymbol z_e$, encoding their inherent, time-invariant properties. The detailed formulation is provided in Appendix~\ref{sec:static-embedding-details}.

\subsection{Sparsity-aware Dynamic Embedding} \label{sec:dynamic-embedding}

During a specific time slot $T_j$, most road segments have few known microscopic weights, presenting segment-layer sparsity. The sparsity-aware dynamic embedding module addresses this by adaptively balancing two information sources:
(1) the \textit{aggregated known weights} of the segment during the time slot, and
(2) the \textit{static embedding vector} $\boldsymbol z_e$ learned in Section~\ref{sec:embedding-module}, which captures inherent segment properties.

We first aggregate known weights $\mathbb W_e^{T_j}$ using learnable Fourier features and a Transformer encoder~\cite{DBLP:conf/nips/VaswaniSPUJGKP17}, yielding $\mathrm{Agg}_e^{T_j} \in \mathbb R^d$. To incorporate sparsity-awareness, we encode the number of known weights $|\mathbb W_e^{T_j}|$ into a hidden vector $\Psi_e^{T_j}$ using learnable Fourier features, then employ a gate mechanism~\cite{DBLP:journals/corr/ChungGCB14} to balance the two sources:
\begin{equation}
\begin{split}
\boldsymbol f_e^{T_j} &= \sigma({\Psi_e^{T_j}}^\top \boldsymbol W_f \mathrm{Agg}_e^{T_j} + {\Psi_e^{T_j}}^\top \boldsymbol U_f \boldsymbol z_e + \boldsymbol b_f),\\
\hat{\boldsymbol{h}}_e^{T_j} &= \mathrm{tanh}(\boldsymbol W_h \mathrm{Agg}_e^{T_j} + \boldsymbol U_h(\boldsymbol f_e^{T_j}\odot \boldsymbol z_e) + \boldsymbol b_h), \\
\boldsymbol{h}_e^{T_j} &= (1-\boldsymbol f_e^{T_j})\odot \boldsymbol z_e + \boldsymbol f_e^{T_j} \odot \hat{\boldsymbol{h}}_e^{T_j},
\end{split}
\label{eq:segment-hidden-state}
\end{equation}
where $\boldsymbol f_e^{T_j}$ is a forget gate controlled by the sparsity encoding, $\hat{\boldsymbol{h}}_e^{T_j}$ is a candidate hidden vector combining both sources, and $\boldsymbol{h}_e^{T_j}$ is the dynamic embedding vector of segment $e$ during time slot $T_j$. When fewer weights are known (higher sparsity), the gate relies more on the static embedding $\boldsymbol z_e$; when more weights are available, it incorporates more from the aggregated known weights. The detailed formulation is provided in Appendix~\ref{sec:dynamic-embedding-details}.

\subsection{Spatiotemporal U-net with Global Residual Connections} \label{sec:spatiotemporal-unet}

Accurate microscopic weight completion relies heavily on long-range spatial correlations and long-term temporal dependencies. To effectively model such correlations and address network-layer sparsity, we propose a spatiotemporal U-net with global residual connections, building upon the U-net architecture~\cite{DBLP:conf/miccai/RonnebergerFB15,DBLP:conf/nips/HoJA20}.

The spatiotemporal U-net consists of spatiotemporal residual (STRes) blocks and down/up-sampling blocks organized in three components: the left component with $L_\mathrm{res}$ pairs of STRes and down-sampling blocks, the middle component with two STRes blocks, and the right component with $L_\mathrm{res}$ pairs of STRes and up-sampling blocks.

\subsubsection{STRes and Down/Up-Sampling Blocks}

Following UGnet~\cite{DBLP:conf/gis/WenLXWWZL23}, each STRes block captures spatiotemporal correlations through temporal convolution (TCN~\cite{lea2016temporal}) followed by spatial convolution (GCN), with a residual connection. The input to the first STRes block is the collection of dynamic embeddings $\boldsymbol{h}_e^{T_j}$ from all segments across all time slots, forming $\boldsymbol H \in \mathbb R^{|\mathcal E| \times H \times d}$. Subsequent blocks take outputs from previous blocks.

The down-sampling blocks halve the temporal dimension while doubling the feature dimension, enabling effective modeling of long-term temporal correlations. The up-sampling blocks perform the reverse operation. This multi-scale structure avoids excessive stacking of STRes blocks, which can cause vanishing gradients. Detailed formulations are provided in Appendix~\ref{sec:unet-details}.

\subsubsection{Global Residual Connections}

In most U-net architectures, residual connections exist from the left component to the right component to facilitate effective training. We introduce \textit{global residual connections} to enhance the modeling of long-range spatial correlations and further address network-layer sparsity.

We first partition the edge set $\mathcal E$ of road network $\mathcal G$ into non-intersecting clusters using the Louvain algorithm~\cite{DBLP:journals/corr/abs-2311-06047}:
\begin{equation}
\mathcal E \rightarrow \{\mathcal E_1, \mathcal E_2, \dots, \mathcal E_{N_\mathrm{cls}}\},
\end{equation}
where $\mathcal E_j$ is a set of edges belonging to the $j$-th cluster and $N_\mathrm{cls}$ is the number of clusters.

For each STRes block in the left component, we perform cluster-level mean pooling on its input $\boldsymbol H$ to obtain a clustered hidden state for the $j$-th cluster:
\begin{equation}
\boldsymbol H_{\mathrm{cls},j} = \frac{\sum_{e_i^{(j)}\in \mathcal E_j} \boldsymbol H_{e_i^{(j)}}}{|\mathcal E_j|} \in \mathbb R^{N_\mathrm{in} \times d_\mathrm{in}}.
\end{equation}

We then expand the clustered representations back to segment-level by fetching each segment's cluster representation with positional encoding:
\begin{equation}
\boldsymbol H_{\mathrm{exp}, e_i} = \boldsymbol H_{\mathrm{cls}, j} + \mathrm{PE}(i),
\label{eq:fetched-hidden-state-from-cluster}
\end{equation}
where $e_i$ belongs to the $j$-th cluster and $\mathrm{PE}$ is positional encoding~\cite{DBLP:conf/nips/VaswaniSPUJGKP17} to distinguish different segments within a cluster. The expanded representations $\boldsymbol H_\mathrm{exp} \in \mathbb R^{|\mathcal E| \times N_\mathrm{in} \times d_\mathrm{in}}$ are concatenated with $\boldsymbol H$ along the feature dimension as residual input to the corresponding STRes block in the right component.

By establishing residual connections based on graph clustering, the spatiotemporal U-net extracts long-range spatial correlations without excessive stacking of GCN layers, which can cause over-smoothing.

\subsection{Learnable Gaussian Mixture Model}

To address the challenge of microscopic weight representation, we adopt learnable Gaussian mixture models (GMMs). Unlike histograms whose accuracy is severely affected by heavy tails, and diffusion models that lack closed-form representations, GMMs provide closed-form distributions that flexibly capture complex traffic conditons including heavy tails and multiple clusters.

From the output of the spatiotemporal U-net $\boldsymbol {\bar H} \in \mathbb R^{|\mathcal E| \times H \times d}$, we estimate the GMM for each segment $e$ and time slot $T_j$ by applying three prediction heads on $\boldsymbol {\bar H}_e^{T_j}$:
\begin{equation}
\begin{split}
\langle \phi_1, \phi_2, \dots, \phi_K \rangle &= \mathrm{softmax}(\boldsymbol {\bar H}_e^{T_j} \boldsymbol W_\phi + \boldsymbol b_\phi), \\
\langle \mu_1, \mu_2, \dots, \mu_K \rangle &= \mathrm{ReLU}(\boldsymbol {\bar H}_e^{T_j} \boldsymbol W_\mu + \boldsymbol b_\mu), \\
\langle \sigma_1, \sigma_2, \dots, \sigma_K \rangle &= \exp(\boldsymbol {\bar H}_e^{T_j} \boldsymbol W_\sigma + \boldsymbol b_\sigma), \\
p(w_e^{T_j}) &= \sum_{k=1}^K \phi_k \mathcal N(w_e^{T_j}|\mu_k,\sigma_k),
\end{split}
\label{eq:gmm-estimation}
\end{equation}
where $\boldsymbol W_\phi, \boldsymbol W_\mu, \boldsymbol W_\sigma \in \mathbb R^{d\times K}$ and $\boldsymbol b_\phi, \boldsymbol b_\mu, \boldsymbol b_\sigma \in \mathbb R^K$ are learnable parameters, and $K$ is the number of Gaussian components.

To train the model, we maximize the likelihood of $p(w_e^{T_j})$ for all observed microscopic weights. Given a ground truth weight $w$, the loss is:
\begin{equation}
\begin{split}
\mathcal L_\mathrm{MLE} = \sum_{k=1}^K \phi_k\exp( &-({(w-\mu_k)}^2/2{\sigma_k}^2) \\
&- \log(\sigma_k) - \log(\sqrt{2\pi})).
\end{split}
\end{equation}
While inference typically completes weights only for the most recent time slot $T_H$, we train on all time slots in $\mathcal T$ to enhance model effectiveness.

\section{Experiments}
% To evaluate the proposed method's effectiveness at microscopic weight completion, we use two real-world microscopic weight datasets, and compare the proposed method against several existing state-of-the-art methods on the datasets.

\subsection{Datasets}
We utilize two microscopic weight datasets referred to as \textbf{HTT} and \textbf{CD}.
Both datasets are organized into records of vehicle-level travel speeds (m/s) on road segments.

The HTT dataset was published at the KDD Cup 2017\footnote{https://www.kdd.org/kdd2017/announcements/view/announcing-kdd-cup-2017-highway-tollgates-traffic-flow-prediction} for a highway tollgate traffic flow prediction competition. It contains travel times of 24 road segments recorded by loop detectors located in a highway tollgate network. The data covers 19/07/2016 to 31/10/2016.
We calculate travel speeds of road segments utilizing the travel times and road segment lengths in the datasets.

The CD dataset was released by Didi\footnote{https://gaia.didichuxing.com/} and consists of GPS trajectories of taxis operating in Chengdu during 01/10/2018 to 01/12/2018. We fetch the road network of Chengdu from OpenStreetMap\footnote{https://www.openstreetmap.org/} and map-match~\cite{DBLP:journals/gis/YangG18} the trajectories onto the road network.
The travel speeds of road segments are then calculated based on the map-matched trajectory points.
To have a sufficient number of ground truth weights for evaluation, we select road segments in the central region of Chengdu and use the largest connected subgraph consisting of 194 road segments.

For both datasets, we divide a day into 96 15-minute time slots. This allows us to form a set of time-varying microscopic weights for each segment and time slot.
The original missing rate of a dataset is defined the number of empty sets divided by the total number of sets in the dataset. Given the above settings, the original missing rates of HTT and CD are 31.918\% and 33.766\%, respectively.

To ensure that the completed representation of weight distributions can be evaluated, we follow the practice of GCWC and use target missing rates to process the input weights to models.
Specifically, for each time slot, given a target missing rate $r$, we remove all known weights of randomly selected road segments so that there are at least a fraction $r$ of road segments miss known weights.
The processed sets of weights are given as inputs to the models, and the unprocessed sets are used as the ground truth. In our experiments, four target missing rates are used: $r=50\%$, $60\%$, $70\%$, and $80\%$.

\subsection{Comparison Methods}
The following methods are direct comparison with the proposed method on microscopic weight completion.

\begin{itemize}[leftmargin=*]
  \item Historical Average (\textbf{HA}): gathers all historical weights for the target road segment and estimates the density. We use two variants: \textbf{HA-Hist} and \textbf{HA-GMM} to estimate the distributions as histograms and Gaussian mixture models, respectively.
  \item \textbf{GCWC}~\cite{DBLP:conf/icde/HuG0J19}: utilizes GCN-based graph pooling to complete the stochastic weights of road segments.
  \item \textbf{SSTGCN}~\cite{DBLP:conf/gis/Muniz-CuzaHZP022}: combines TCNs and GCNs to model spatiotemporal correlations between road segments.
  \item \textbf{ConGC}~\cite{DBLP:conf/sdm/HanCGMM022}: introduces a contextual graph for incorporating contextual features of road segments.
  \item \textbf{Nuhuo}~\cite{yuan2024nuhuo}: partitions the road network into regions and uses a two-level encoding scheme with global and local encoders. We also construct a variant \textbf{Nuhuo-GMM} by modifying the original output distributions from histograms to Gaussian mixture models.
\end{itemize}

We also include two state-of-the-art macroscopic weight completion methods with uncertainty consideration for comparison.
Both methods share the technical similarity with microscoic weight completion: they estimate weight distributions.
Do note that since they focus on macroscopic weight, the distribution estimation design is to take the uncertainty of statistics into consideration. This is fundamentally different from methods focus on microscopic weights.

\begin{itemize}[leftmargin=*]
  \item \textbf{STGNF}~\cite{an2024spatio}: combines conditional normalizing flows with spatiotemporal graph learning to predict the probability distribution of macroscopic weights.
  \item \textbf{PriSTI}~\cite{liu2023pristi}: employs conditional diffusion probabilistic models with spatiotemporal attention to estimate the probability distribution of macroscopic weights.
\end{itemize}

\subsection{Settings}
For both datasets, we split the time periods by 8:1:1 to create the training, validation, and testing sets. Models are trained on the training set and evaluated on the testing set. The validation set is used for hyper-parameter tuning and for implementing early-stopping techniques.

To compare the accuracy of various types of distributions fairly, we employ two metrics: likelihood~\cite{shao2003mathematical} and Continuous Ranked Probability Score (CRPS)~\cite{matheson1976scoring}.
Given an estimated weight distribution $p$ and a ground truth weight $w$, the likelihood is directly calculated as the probability of $p$ at observation $w$, and CRPS is calculated as $\mathrm{CRPS}(p, x) = \int_\mathbb R (p(z) - \mathbb I(x\leq z))^2 dz$, where $\mathbb I$ an indicator function which equals one if $x\leq z$, and equals zero otherwise.
These metrics assess how well the estimated distributions align with the recorded microscopic weights.
A higher likelihood and a lower CRPS indicate better performance in the experiment.
Note that likelihood cannot be calculated on methods that do not provide closed-form distributions, like PriSTI.

The proposed method is implemented using PyTorch~\cite{DBLP:conf/nips/PaszkeGMLBCKLGA19}. The five key hyper-parameters of the proposed method and their optimal values are $H=16$, $L_\mathrm{agg}=2$, $L_\mathrm{res}=2$, $d=128$, and $K=4$. The effectiveness of these hyper-parameters are also discussed in Appendix~\ref{sec:hyper-parameters}. 
For the comparison methods, we set their hyper-parameters to the optimal values indicated in their respective papers, and use 8 bins for histograms.
We utilize the contextual information in datasets, including lengths and levels of segments, to build the contextual graph for ConGC.
Each experiment is repeated 5 times, and the mean and variance of the results are reported.

\begin{table*}[t]
\centering
\begin{threeparttable}
\resizebox{1.0\linewidth}{!}{
\begin{tabular}{c|c|cccc|cccc}
\toprule
\multirow{2}{*}{Datasets} & \multirow{2}{*}{Methods} & \multicolumn{4}{c|}{Likelihood (\%) $\uparrow$} & \multicolumn{4}{c}{CRPS $\downarrow$} \\ \cline{3-10} 
 &  & $r=50\%$ & $r=60\%$ & $r=70\%$ & $r=80\%$ & $r=50\%$ & $r=60\%$ & $r=70\%$ & $r=80\%$ \\
 \midrule
 \multirow{10}{*}{HTT} 
 & HA-Hist & 7.290$\pm$0.00 & 7.300$\pm$0.01 & 7.288$\pm$0.01 & 7.288$\pm$0.00 & 
    2.344$\pm$0.00 & 2.345$\pm$0.00 & 2.346$\pm$0.00 & 2.346$\pm$0.00 \\
 & GCWC & 9.343$\pm$0.10 & 9.167$\pm$0.06 & 8.960$\pm$0.05 & 8.576$\pm$0.10 & 
    2.134$\pm$0.01 & 2.160$\pm$0.01 & 2.180$\pm$0.03 & 2.226$\pm$0.02 \\
 & SSTGCN & 10.079$\pm$0.16 & 9.762$\pm$0.15 & 9.362$\pm$0.12 & 8.623$\pm$0.13 & 
    2.111$\pm$0.02 & 2.146$\pm$0.03 & 2.189$\pm$0.02 & 2.229$\pm$0.03 \\
 & ConGC & 10.243$\pm$0.20 & 9.981$\pm$0.23 & 9.717$\pm$0.25 & 9.233$\pm$0.25 & 
    2.098$\pm$0.03 & 2.119$\pm$0.04 & 2.137$\pm$0.04 & 2.183$\pm$0.03 \\
 & Nuhuo & 11.203$\pm$0.20 & 10.564$\pm$0.23 & 10.059$\pm$0.13 & 9.481$\pm$0.12 &
    2.078$\pm$0.01 & 2.104$\pm$0.02 & 2.128$\pm$0.01 & 2.173$\pm$0.01 \\
 \cmidrule{2-10}
 & HA-GMM & 6.989$\pm$0.03 & 6.959$\pm$0.03 & 6.974$\pm$0.02 & 6.991$\pm$0.04 & 
    2.364$\pm$0.00 & 2.367$\pm$0.00 & 2.365$\pm$0.00 & 2.365$\pm$0.00 \\
 & STGNF & 12.195$\pm$0.15 & 11.268$\pm$0.17 & 10.445$\pm$0.10 & 9.083$\pm$0.19 &
    2.087$\pm$0.02 & 2.112$\pm$0.01 & 2.183$\pm$0.02 & 2.283$\pm$0.02 \\
 & PriSTI & - & - & - & - &
    2.055$\pm$0.04 & 2.082$\pm$0.04 & 2.115$\pm$0.03 & 2.169$\pm$0.03 \\
 & Nuhuo-GMM & \underline{12.915$\pm$0.15} & \underline{12.446$\pm$0.13} & \underline{11.724$\pm$0.13} & \underline{10.779$\pm$0.10} & 
    \underline{2.032$\pm$0.01} & \underline{2.062$\pm$0.02} & \underline{2.101$\pm$0.02} & \underline{2.165$\pm$0.02} \\
 \cmidrule{2-10}
 & \textbf{DiSGMM (ours)} & \textbf{14.745}$\pm$\textbf{0.12} & \textbf{13.964}$\pm$\textbf{0.22} & \textbf{12.948}$\pm$\textbf{0.23} & \textbf{11.619}$\pm$\textbf{0.23} & 
  \textbf{1.936}$\pm$\textbf{0.01} & \textbf{1.972}$\pm$\textbf{0.01} & \textbf{2.004}$\pm$\textbf{0.02} & \textbf{2.067}$\pm$\textbf{0.01} \\
 \midrule
 
 \multirow{10}{*}{CD} 
 & HA-Hist & 6.444$\pm$0.00 & 6.438$\pm$0.00 & 6.436$\pm$0.01 & 6.442$\pm$0.02 & 
    2.628$\pm$0.00 & 2.630$\pm$0.00 & 2.628$\pm$0.00 & 2.630$\pm$0.00 \\
 & GCWC & 7.704$\pm$0.04 & 7.633$\pm$0.05 & 7.508$\pm$0.12 & 7.440$\pm$0.07 & 
    2.365$\pm$0.02 & 2.378$\pm$0.01 & 2.381$\pm$0.02 & 2.388$\pm$0.02 \\
 & SSTGCN & 8.663$\pm$0.07 & 8.321$\pm$0.05 & 7.828$\pm$0.10 & 7.224$\pm$0.07 & 
    2.226$\pm$0.02 & 2.321$\pm$0.03 & 2.447$\pm$0.04 & 2.640$\pm$0.03 \\
 & ConGC & 8.466$\pm$0.17 & 8.353$\pm$0.14 & 8.357$\pm$0.27 & 8.158$\pm$0.16 & 
    2.293$\pm$0.02 & 2.333$\pm$0.04 & 2.311$\pm$0.04 & 2.351$\pm$0.04 \\
 & Nuhuo & 9.559$\pm$0.25 & 9.107$\pm$0.19 & 8.705$\pm$0.23 & 8.444$\pm$0.13 &
    2.246$\pm$0.01 & 2.287$\pm$0.02 & 2.304$\pm$0.04 & 2.343$\pm$0.02 \\
 \cmidrule{2-10}
 & HA-GMM & 7.469$\pm$0.02 & 7.421$\pm$0.01 & 7.463$\pm$0.02 & 7.485$\pm$0.00 & 
    2.519$\pm$0.01 & 2.525$\pm$0.00 & 2.518$\pm$0.01 & 2.517$\pm$0.00 \\
 & STGNF & \underline{13.957$\pm$0.18} & 11.925$\pm$0.53 & 10.217$\pm$0.36 & 8.555$\pm$0.15 &
    \underline{2.109$\pm$0.04} & 2.195$\pm$0.06 & 2.371$\pm$0.06 & 2.582$\pm$0.06 \\
 & PriSTI & - & - & - & - &
    2.135$\pm$0.05 & 2.185$\pm$0.05 & 2.268$\pm$0.06 & 2.315$\pm$0.06 \\
 & Nuhuo-GMM & 13.905$\pm$0.22 & \underline{12.878$\pm$0.23} & \underline{11.690$\pm$0.24} & \underline{10.368$\pm$0.15} & 
    2.116$\pm$0.02 & \underline{2.165$\pm$0.03} & \underline{2.232$\pm$0.03} & \underline{2.273$\pm$0.03} \\
 \cmidrule{2-10}
 & \textbf{DiSGMM (ours)} & \textbf{16.388}$\pm$\textbf{0.13} & \textbf{15.027}$\pm$\textbf{0.14} & \textbf{13.403}$\pm$\textbf{0.22} & \textbf{11.835}$\pm$\textbf{0.15} & 
  \textbf{1.996}$\pm$\textbf{0.03} & \textbf{2.043}$\pm$\textbf{0.02} & \textbf{2.114}$\pm$\textbf{0.05} & \textbf{2.145}$\pm$\textbf{0.03} \\
 \bottomrule
\end{tabular}}
\begin{tablenotes}\footnotesize
    \item[]{\textbf{Bold} denotes the best result, and \underline{underline} denotes the second-best result.}
\end{tablenotes}
\end{threeparttable}

\vspace{-6pt}
\caption{Overall performance of methods.}
\label{table:overall-performance}
\end{table*}

\begin{table*}[t]
\centering
\begin{threeparttable}
\resizebox{1.0\linewidth}{!}{
\begin{tabular}{c|cccc|cccc}
\toprule
\multirow{2}{*}{Methods} & \multicolumn{4}{c|}{Likelihood (\%) $\uparrow$} & \multicolumn{4}{c}{CRPS $\downarrow$} \\ \cline{2-9} 
 & $r=50\%$ & $r=60\%$ & $r=70\%$ & $r=80\%$ & $r=50\%$ & $r=60\%$ & $r=70\%$ & $r=80\%$ \\
 \midrule
 -Sparsity & \underline{14.696$\pm$0.30} & 13.591$\pm$0.28 & \underline{12.603$\pm$0.22} & 11.198$\pm$0.14 & 
   \underline{1.940$\pm$0.02} & 1.984$\pm$0.01 & \underline{2.010$\pm$0.01} & \underline{2.071$\pm$0.03} \\
 -Static & 13.420$\pm$0.15 & 13.040$\pm$0.21 & 12.018$\pm$0.37 & 10.813$\pm$0.31 & 
   1.962$\pm$0.01 & 2.003$\pm$0.02 & 2.058$\pm$0.03 & 2.098$\pm$0.04 \\
 -DS & 13.026$\pm$0.07 & 12.716$\pm$0.05 & 11.436$\pm$0.22 & 10.283$\pm$0.18 & 
   1.991$\pm$0.02 & 2.015$\pm$0.02 & 2.087$\pm$0.02 & 2.171$\pm$0.02  \\
 -Global Res & 14.565$\pm$0.20 & \underline{13.643$\pm$0.18} & 12.410$\pm$0.23 & \underline{11.462$\pm$0.15} & 
   1.946$\pm$0.02 & \underline{1.981$\pm$0.01} & 2.018$\pm$0.03 & 2.074$\pm$0.03 \\
 \textbf{DiSGMM (full)} & \textbf{14.745$\pm$0.12} & \textbf{13.964$\pm$0.22} & \textbf{12.948$\pm$0.23} & \textbf{11.619$\pm$0.23} & 
   \textbf{1.936$\pm$0.01} & \textbf{1.972$\pm$0.01} & \textbf{2.004$\pm$0.02} & \textbf{2.067$\pm$0.01} \\
 \bottomrule
\end{tabular}}
\begin{tablenotes}\footnotesize
    \item[]{\textbf{Bold} denotes the best result, and \underline{underline} denotes the second-best result.}
\end{tablenotes}
\end{threeparttable}

\vspace{-6pt}
\caption{Performance of DiSGMM's variants and the full model.}
\label{table:components}
\end{table*}

\subsection{Performance Comparison}
Table~\ref{table:overall-performance} presents the overall performance of the methods. The estimated weight distributions are also visualized in Appendix~\ref{sec:case-study}.

We have the following observations.
First, among the histogram-based microscopic weight completion methods, Nuhuo achieves the best performance by leveraging a two-level encoding scheme that captures both global and local spatial patterns.
GCWC and SSTGCN, which model spatial or spatiotemporal correlations, outperform the simple historical average baseline HA-Hist. ConGC further improves accuracy by incorporating contextual features of road segments.

Second, methods that model microscopic weights as GMMs generally achieve higher accuracy than histogram-based methods.
Histograms are limited by their predefined range and number of bins, making it difficult to accurately capture the underlying weight distributions.
In contrast, GMMs can adapt to complex traffic conditions by learning appropriate means, variances, and mixture weights.
This advantage is reflected in the performance gap between Nuhuo and Nuhuo-GMM, where the GMM variant consistently outperforms its histogram counterpart.

Third, the macroscopic weight completion methods, STGNF and PriSTI, show mixed results.
STGNF is designed for traffic prediction rather than completion, and thus does not explicitly consider either layer of sparsity; its performance degrades notably at high missing rates.
PriSTI models only macroscopic weights (aggregated statistics) and does not account for segment-level sparsity, which limits its effectiveness in the microscopic setting.

Fourth, DiSGMM achieves the best performance among all methods.
While Nuhuo and Nuhuo-GMM address network-layer sparsity through their encoding schemes, they do not explicitly model segment-level sparsity.
DiSGMM addresses both layers of sparsity: it calculates sparsity-aware embeddings that effectively capture sparsity at the segment level, and employs a spatiotemporal U-net architecture to capture long-range correlations across the network.
This comprehensive treatment of sparsity enables DiSGMM to achieve superior accuracy in microscopic weight completion.

\subsection{Ablation Study}
To assess the effectiveness of the components of DiSGMM, we compare the performance of the complete method against four variants:
\begin{itemize}[leftmargin=*]
    \item -\textit{Sparsity}: removes sparsity-awareness from the gated mechanism in Equation~\ref{eq:segment-hidden-state}. Specifically, makes $\boldsymbol f_e^{T_j} = \sigma(\boldsymbol W_f' \mathrm{Agg}_e^{T_j} + \boldsymbol U_f' \boldsymbol z_e + \boldsymbol b_f)$.
    \item -\textit{Static}: removes the static road segment embedding module and uses instead the vanilla node2vec for learning road segment embeddings.
    \item -\textit{DS}: removes both sparsity-awareness and static embeddings, and uses $\mathrm{Agg}(\mathbb W_e^{T_j})$ from as the hidden state $\boldsymbol{h}_e^{T_j}$.
    \item -\textit{Global Res}: removes the global residual connections.
\end{itemize}

The performance of these variants on the HTT dataset is reported in Table~\ref{table:components}.
First, comparing the variants \textit{-Sparsity}, \textit{-Static}, and \textit{-DS} with the full model, we find that the sparsity-awareness of the gated mechanism in Equation~\ref{eq:segment-hidden-state} and the static road segment embedding module both improve the performance of DiSGMM. Removing either or both of these designs leads to a drop in the completion accuracy.
Second, comparing the variant \textit{-Global Res} with the full model, we observe that the global residual connections in the proposed spatiotemporal U-net are effective. Removing these connections impacts the model's ability to capture long-range spatial correlations negatively and results in worse performance.

\section{Conclusion}
We propose DiSGMM, a novel method for time-varying microscopic weight completion on road networks that addresses two unique challenges.
For the two layers of sparsity, DiSGMM incorporates sparsity-aware embeddings that effectively leverage sparse observations alongside learned segment properties, and a spatiotemporal U-net with global residual connections that captures long-range correlations.
For representation design, DiSGMM adopts learnable Gaussian mixture models, providing closed-form distributions that flexibly capture complex traffic conditions.
Extensive experiments on two real-world datasets demonstrate that DiSGMM consistently outperforms state-of-the-art methods.

\section*{Ethical Statement}

There are no ethical issues.

\paragraph{Use of large language models (LLMs).} The use of LLMs in this work is restricted to two aspects: 1) For proofreading purposes after the drafting of this paper is complete; 2) For writing code to convert numerical experimental results into illustration plots.

\section*{Acknowledgments}

\bibliographystyle{named}
\bibliography{main}

\begin{thebibliography}{}

\bibitem[\protect\citeauthoryear{Abou~Elassad \bgroup \em et al.\egroup
  }{2020}]{abou2020application}
Zouhair~Elamrani Abou~Elassad, Hajar Mousannif, Hassan Al~Moatassime, and Aimad
  Karkouch.
\newblock The application of machine learning techniques for driving behavior
  analysis: A conceptual framework and a systematic literature review.
\newblock {\em Eng. Appl. Artif. Intell.}, 87:103312, 2020.

\bibitem[\protect\citeauthoryear{Akhtar and Moridpour}{2021}]{akhtar2021review}
Mahmuda Akhtar and Sara Moridpour.
\newblock A review of traffic congestion prediction using artificial
  intelligence.
\newblock {\em J. Adv. Transp.}, 2021(1):8878011, 2021.

\bibitem[\protect\citeauthoryear{An \bgroup \em et al.\egroup
  }{2024}]{an2024spatio}
Yang An, Zhibin Li, Wei Liu, Haoliang Sun, Meng Chen, Wenpeng Lu, and Yongshun
  Gong.
\newblock Spatio-temporal graph normalizing flow for probabilistic traffic
  prediction.
\newblock In {\em CIKM}, pages 45--55, 2024.

\bibitem[\protect\citeauthoryear{Blondel \bgroup \em et al.\egroup
  }{2008}]{DBLP:journals/corr/abs-2311-06047}
Vincent~D Blondel, Jean-Loup Guillaume, Renaud Lambiotte, and Etienne Lefebvre.
\newblock Fast unfolding of communities in large networks.
\newblock {\em J. Stat. Mech. Theory Exp.}, 2008(10):P10008, 2008.

\bibitem[\protect\citeauthoryear{Cao \bgroup \em et al.\egroup
  }{2018}]{wei2018brits}
Wei Cao, Dong Wang, Jian Li, Hao Zhou, Lei Li, and Yitan Li.
\newblock {BRITS:} bidirectional recurrent imputation for time series.
\newblock In {\em NeurIPS}, pages 6776--6786, 2018.

\bibitem[\protect\citeauthoryear{Chen \bgroup \em et al.\egroup
  }{2022}]{chen2021low}
Xinyu Chen, Mengying Lei, Nicolas Saunier, and Lijun Sun.
\newblock Low-rank autoregressive tensor completion for spatiotemporal traffic
  data imputation.
\newblock {\em {IEEE} Trans. Intell. Transp. Syst.}, 23(8):12301--12310, 2022.

\bibitem[\protect\citeauthoryear{Chung \bgroup \em et al.\egroup
  }{2014}]{DBLP:journals/corr/ChungGCB14}
Junyoung Chung, Caglar Gulcehre, KyungHyun Cho, and Yoshua Bengio.
\newblock Empirical evaluation of gated recurrent neural networks on sequence
  modeling.
\newblock {\em arXiv preprint arXiv:1412.3555}, 2014.

\bibitem[\protect\citeauthoryear{Cini \bgroup \em et al.\egroup
  }{2022}]{cini2021filling}
Andrea Cini, Ivan Marisca, and Cesare Alippi.
\newblock Filling the g\_ap\_s: Multivariate time series imputation by graph
  neural networks.
\newblock In {\em ICLR}, 2022.

\bibitem[\protect\citeauthoryear{Fang \bgroup \em et al.\egroup
  }{2022}]{DBLP:conf/kdd/Fang0ZHCGJ22}
Ziquan Fang, Yuntao Du, Xinjun Zhu, Danlei Hu, Lu~Chen, Yunjun Gao, and
  Christian~S. Jensen.
\newblock Spatio-temporal trajectory similarity learning in road networks.
\newblock In {\em SIGKDD}, pages 347--356, 2022.

\bibitem[\protect\citeauthoryear{Gao \bgroup \em et al.\egroup
  }{2015}]{DBLP:journals/tkde/GaoQZ015}
Yunjun Gao, Xu~Qin, Baihua Zheng, and Gang Chen.
\newblock Efficient reverse top-k boolean spatial keyword queries on road
  networks.
\newblock {\em {IEEE} Trans. Knowl. Data Eng.}, 27(5):1205--1218, 2015.

\bibitem[\protect\citeauthoryear{Goodfellow \bgroup \em et al.\egroup
  }{2014}]{goodfellow2014generative}
Ian~J. Goodfellow, Jean Pouget{-}Abadie, Mehdi Mirza, Bing Xu, David
  Warde{-}Farley, Sherjil Ozair, Aaron~C. Courville, and Yoshua Bengio.
\newblock Generative adversarial nets.
\newblock In {\em NeurIPS}, pages 2672--2680, 2014.

\bibitem[\protect\citeauthoryear{Guo \bgroup \em et al.\egroup
  }{2019}]{DBLP:conf/aaai/GuoLFSW19}
Shengnan Guo, Youfang Lin, Ning Feng, Chao Song, and Huaiyu Wan.
\newblock Attention based spatial-temporal graph convolutional networks for
  traffic flow forecasting.
\newblock In {\em AAAI}, pages 922--929, 2019.

\bibitem[\protect\citeauthoryear{Han \bgroup \em et al.\egroup
  }{2022}]{DBLP:conf/sdm/HanCGMM022}
Xiaolin Han, Reynold Cheng, Tobias Grubenmann, Silviu Maniu, Chenhao Ma, and
  Xiaodong Li.
\newblock Leveraging contextual graphs for stochastic weight completion in
  sparse road networks.
\newblock In {\em SDM}, pages 64--72, 2022.

\bibitem[\protect\citeauthoryear{Ho \bgroup \em et al.\egroup
  }{2020a}]{ho2020denoising}
Jonathan Ho, Ajay Jain, and Pieter Abbeel.
\newblock Denoising diffusion probabilistic models.
\newblock {\em NeurIPS}, 33:6840--6851, 2020.

\bibitem[\protect\citeauthoryear{Ho \bgroup \em et al.\egroup
  }{2020b}]{DBLP:conf/nips/HoJA20}
Jonathan Ho, Ajay Jain, and Pieter Abbeel.
\newblock Denoising diffusion probabilistic models.
\newblock In {\em NeurIPS}, 2020.

\bibitem[\protect\citeauthoryear{Hochreiter and
  Schmidhuber}{1997}]{hochreiter1997long}
Sepp Hochreiter and J{\"{u}}rgen Schmidhuber.
\newblock Long short-term memory.
\newblock {\em Neural Comput.}, 9(8):1735--1780, 1997.

\bibitem[\protect\citeauthoryear{Hollander and
  Liu}{2008}]{hollander2008principles}
Yaron Hollander and Ronghui Liu.
\newblock The principles of calibrating traffic microsimulation models.
\newblock {\em Transportation}, 35(3):347--362, 2008.

\bibitem[\protect\citeauthoryear{Hu \bgroup \em et al.\egroup
  }{2019}]{DBLP:conf/icde/HuG0J19}
Jilin Hu, Chenjuan Guo, Bin Yang, and Christian~S. Jensen.
\newblock Stochastic weight completion for road networks using graph
  convolutional networks.
\newblock In {\em ICDE}, pages 1274--1285, 2019.

\bibitem[\protect\citeauthoryear{Jiang and Luo}{2022}]{jiang2022graph}
Weiwei Jiang and Jiayun Luo.
\newblock Graph neural network for traffic forecasting: A survey.
\newblock {\em Expert Syst. Appl.}, 207:117921, 2022.

\bibitem[\protect\citeauthoryear{Jin \bgroup \em et al.\egroup
  }{2023}]{jin2023spatio}
Guangyin Jin, Yuxuan Liang, Yuchen Fang, Zezhi Shao, Jincai Huang, Junbo Zhang,
  and Yu~Zheng.
\newblock Spatio-temporal graph neural networks for predictive learning in
  urban computing: A survey.
\newblock {\em IEEE Trans. Knowl. Data Eng.}, 36(10):5388--5408, 2023.

\bibitem[\protect\citeauthoryear{Kipf and Welling}{2017}]{kipf2016semi}
Thomas~N. Kipf and Max Welling.
\newblock Semi-supervised classification with graph convolutional networks.
\newblock In {\em ICLR}, 2017.

\bibitem[\protect\citeauthoryear{Lea \bgroup \em et al.\egroup
  }{2016}]{lea2016temporal}
Colin Lea, Rene Vidal, Austin Reiter, and Gregory~D Hager.
\newblock Temporal convolutional networks: A unified approach to action
  segmentation.
\newblock In {\em ECCV}, pages 47--54, 2016.

\bibitem[\protect\citeauthoryear{Li \bgroup \em et al.\egroup
  }{2019}]{DBLP:conf/iclr/LiJM19}
Steven~Cheng{-}Xian Li, Bo~Jiang, and Benjamin~M. Marlin.
\newblock Misgan: Learning from incomplete data with generative adversarial
  networks.
\newblock In {\em ICLR}, 2019.

\bibitem[\protect\citeauthoryear{Li \bgroup \em et al.\egroup
  }{2021}]{DBLP:conf/nips/LiSLHB21}
Yang Li, Si~Si, Gang Li, Cho{-}Jui Hsieh, and Samy Bengio.
\newblock Learnable fourier features for multi-dimensional spatial positional
  encoding.
\newblock In {\em NeurIPS}, pages 15816--15829, 2021.

\bibitem[\protect\citeauthoryear{Liu \bgroup \em et al.\egroup
  }{2019}]{liu2019naomi}
Yukai Liu, Rose Yu, Stephan Zheng, Eric Zhan, and Yisong Yue.
\newblock {NAOMI:} non-autoregressive multiresolution sequence imputation.
\newblock In {\em NeurIPS}, pages 11236--11246, 2019.

\bibitem[\protect\citeauthoryear{Liu \bgroup \em et al.\egroup
  }{2023}]{liu2023pristi}
Mingzhe Liu, Han Huang, Hao Feng, Leilei Sun, Bowen Du, and Yanjie Fu.
\newblock Pristi: {A} conditional diffusion framework for spatiotemporal
  imputation.
\newblock In {\em ICDE}, pages 1927--1939, 2023.

\bibitem[\protect\citeauthoryear{Matheson and
  Winkler}{1976}]{matheson1976scoring}
James~E Matheson and Robert~L Winkler.
\newblock Scoring rules for continuous probability distributions.
\newblock {\em Manag. Sci.}, 22(10):1087--1096, 1976.

\bibitem[\protect\citeauthoryear{Mikolov \bgroup \em et al.\egroup
  }{2013}]{DBLP:journals/corr/abs-1301-3781}
Tom{\'{a}}s Mikolov, Kai Chen, Greg Corrado, and Jeffrey Dean.
\newblock Efficient estimation of word representations in vector space.
\newblock In {\em ICLR}, 2013.

\bibitem[\protect\citeauthoryear{Mu{\~{n}}iz{-}Cuza \bgroup \em et al.\egroup
  }{2022}]{DBLP:conf/gis/Muniz-CuzaHZP022}
Carlos~Enrique Mu{\~{n}}iz{-}Cuza, Nguyen Ho, Eleni~Tzirita Zacharatou,
  Torben~Bach Pedersen, and Bin Yang.
\newblock Spatio-temporal graph convolutional network for stochastic traffic
  speed imputation.
\newblock In {\em SIGSPATIAL}, pages 14:1--14:12, 2022.

\bibitem[\protect\citeauthoryear{Nie \bgroup \em et al.\egroup
  }{2024}]{nie2024imputeformer}
Tong Nie, Guoyang Qin, Wei Ma, Yuewen Mei, and Jian Sun.
\newblock Imputeformer: Low rankness-induced transformers for generalizable
  spatiotemporal imputation.
\newblock In {\em SIGKDD}, pages 2260--2271, 2024.

\bibitem[\protect\citeauthoryear{Paszke \bgroup \em et al.\egroup
  }{2019}]{DBLP:conf/nips/PaszkeGMLBCKLGA19}
Adam Paszke, Sam Gross, Francisco Massa, Adam Lerer, James Bradbury, Gregory
  Chanan, Trevor Killeen, Zeming Lin, Natalia Gimelshein, Luca Antiga, Alban
  Desmaison, Andreas K{\"{o}}pf, Edward~Z. Yang, Zachary DeVito, Martin Raison,
  Alykhan Tejani, Sasank Chilamkurthy, Benoit Steiner, Lu~Fang, Junjie Bai, and
  Soumith Chintala.
\newblock Pytorch: An imperative style, high-performance deep learning library.
\newblock In {\em NeurIPS}, pages 8024--8035, 2019.

\bibitem[\protect\citeauthoryear{Perozzi \bgroup \em et al.\egroup
  }{2014}]{DBLP:conf/kdd/PerozziAS14}
Bryan Perozzi, Rami Al{-}Rfou, and Steven Skiena.
\newblock Deepwalk: online learning of social representations.
\newblock In {\em SIGKDD}, pages 701--710, 2014.

\bibitem[\protect\citeauthoryear{Reynolds and
  others}{2009}]{reynolds2009gaussian}
Douglas~A Reynolds et~al.
\newblock Gaussian mixture models.
\newblock {\em Encyclopedia of biometrics}, 741(659-663), 2009.

\bibitem[\protect\citeauthoryear{Ronneberger \bgroup \em et al.\egroup
  }{2015}]{DBLP:conf/miccai/RonnebergerFB15}
Olaf Ronneberger, Philipp Fischer, and Thomas Brox.
\newblock U-net: Convolutional networks for biomedical image segmentation.
\newblock In {\em MICCAI}, volume 9351, pages 234--241, 2015.

\bibitem[\protect\citeauthoryear{Shao}{2003}]{shao2003mathematical}
Jun Shao.
\newblock {\em Mathematical statistics}.
\newblock Springer Science \& Business Media, 2003.

\bibitem[\protect\citeauthoryear{Shukla and Marlin}{2021}]{shukla2021multi}
Satya~Narayan Shukla and Benjamin~M. Marlin.
\newblock Multi-time attention networks for irregularly sampled time series.
\newblock In {\em ICLR}, 2021.

\bibitem[\protect\citeauthoryear{Song \bgroup \em et al.\egroup
  }{2019}]{10.1145/3278607}
Qingquan Song, Hancheng Ge, James Caverlee, and Xia Hu.
\newblock Tensor completion algorithms in big data analytics.
\newblock {\em {ACM} Trans. Knowl. Discov. Data}, 13(1):6:1--6:48, 2019.

\bibitem[\protect\citeauthoryear{Tashiro \bgroup \em et al.\egroup
  }{2021}]{tashiro2021csdi}
Yusuke Tashiro, Jiaming Song, Yang Song, and Stefano Ermon.
\newblock {CSDI:} conditional score-based diffusion models for probabilistic
  time series imputation.
\newblock In {\em NeurIPS}, pages 24804--24816, 2021.

\bibitem[\protect\citeauthoryear{Tedjopurnomo \bgroup \em et al.\egroup
  }{2020}]{tedjopurnomo2020survey}
David~Alexander Tedjopurnomo, Zhifeng Bao, Baihua Zheng, Farhana~Murtaza
  Choudhury, and Alex~Kai Qin.
\newblock A survey on modern deep neural network for traffic prediction:
  Trends, methods and challenges.
\newblock {\em IEEE Trans. Knowl. Data Eng.}, 34(4):1544--1561, 2020.

\bibitem[\protect\citeauthoryear{Vaswani \bgroup \em et al.\egroup
  }{2017}]{DBLP:conf/nips/VaswaniSPUJGKP17}
Ashish Vaswani, Noam Shazeer, Niki Parmar, Jakob Uszkoreit, Llion Jones,
  Aidan~N. Gomez, Lukasz Kaiser, and Illia Polosukhin.
\newblock Attention is all you need.
\newblock In {\em NeurIPS}, pages 5998--6008, 2017.

\bibitem[\protect\citeauthoryear{Wei \bgroup \em et al.\egroup
  }{2024}]{DBLP:journals/kbs/WeiLGLZJWW24}
Tonglong Wei, Youfang Lin, Shengnan Guo, Yan Lin, Yiji Zhao, Xiyuan Jin, Zhihao
  Wu, and Huaiyu Wan.
\newblock Inductive and adaptive graph convolution networks equipped with
  constraint task for spatial-temporal traffic data kriging.
\newblock {\em Knowl. Based Syst.}, 284:111325, 2024.

\bibitem[\protect\citeauthoryear{Wen \bgroup \em et al.\egroup
  }{2023a}]{wen2023diffstg}
Haomin Wen, Youfang Lin, Yutong Xia, Huaiyu Wan, Qingsong Wen, Roger
  Zimmermann, and Yuxuan Liang.
\newblock Diffstg: Probabilistic spatio-temporal graph forecasting with
  denoising diffusion models.
\newblock In {\em SIGSPATIAL}, pages 1--12, 2023.

\bibitem[\protect\citeauthoryear{Wen \bgroup \em et al.\egroup
  }{2023b}]{DBLP:conf/gis/WenLXWWZL23}
Haomin Wen, Youfang Lin, Yutong Xia, Huaiyu Wan, Qingsong Wen, Roger
  Zimmermann, and Yuxuan Liang.
\newblock Diffstg: Probabilistic spatio-temporal graph forecasting with
  denoising diffusion models.
\newblock In {\em SIGSPATIAL}, pages 60:1--60:12, 2023.

\bibitem[\protect\citeauthoryear{Wu \bgroup \em et al.\egroup
  }{2021}]{wu2021inductive}
Yuankai Wu, Dingyi Zhuang, Aur{\'{e}}lie Labbe, and Lijun Sun.
\newblock Inductive graph neural networks for spatiotemporal kriging.
\newblock In {\em AAAI}, pages 4478--4485, 2021.

\bibitem[\protect\citeauthoryear{Yang and
  Gid{\'{o}}falvi}{2018}]{DBLP:journals/gis/YangG18}
Can Yang and Gy{\"{o}}z{\"{o}} Gid{\'{o}}falvi.
\newblock Fast map matching, an algorithm integrating hidden markov model with
  precomputation.
\newblock {\em Int. J. Geogr. Inf. Sci.}, 32(3):547--570, 2018.

\bibitem[\protect\citeauthoryear{Yoon \bgroup \em et al.\egroup
  }{2018}]{yoon2018gain}
Jinsung Yoon, James Jordon, and Mihaela van~der Schaar.
\newblock {GAIN:} missing data imputation using generative adversarial nets.
\newblock In {\em ICML}, volume~80, pages 5675--5684, 2018.

\bibitem[\protect\citeauthoryear{Yuan \bgroup \em et al.\egroup
  }{2024}]{yuan2024nuhuo}
Haitao Yuan, Gao Cong, and Guoliang Li.
\newblock Nuhuo: An effective estimation model for traffic speed histogram
  imputation on a road network.
\newblock {\em Proc. VLDB Endow.}, 17(7):1605--1617, 2024.

\end{thebibliography}

\clearpage
\appendix
\section{Effectiveness of Hyper-parameters}
\label{sec:hyper-parameters}

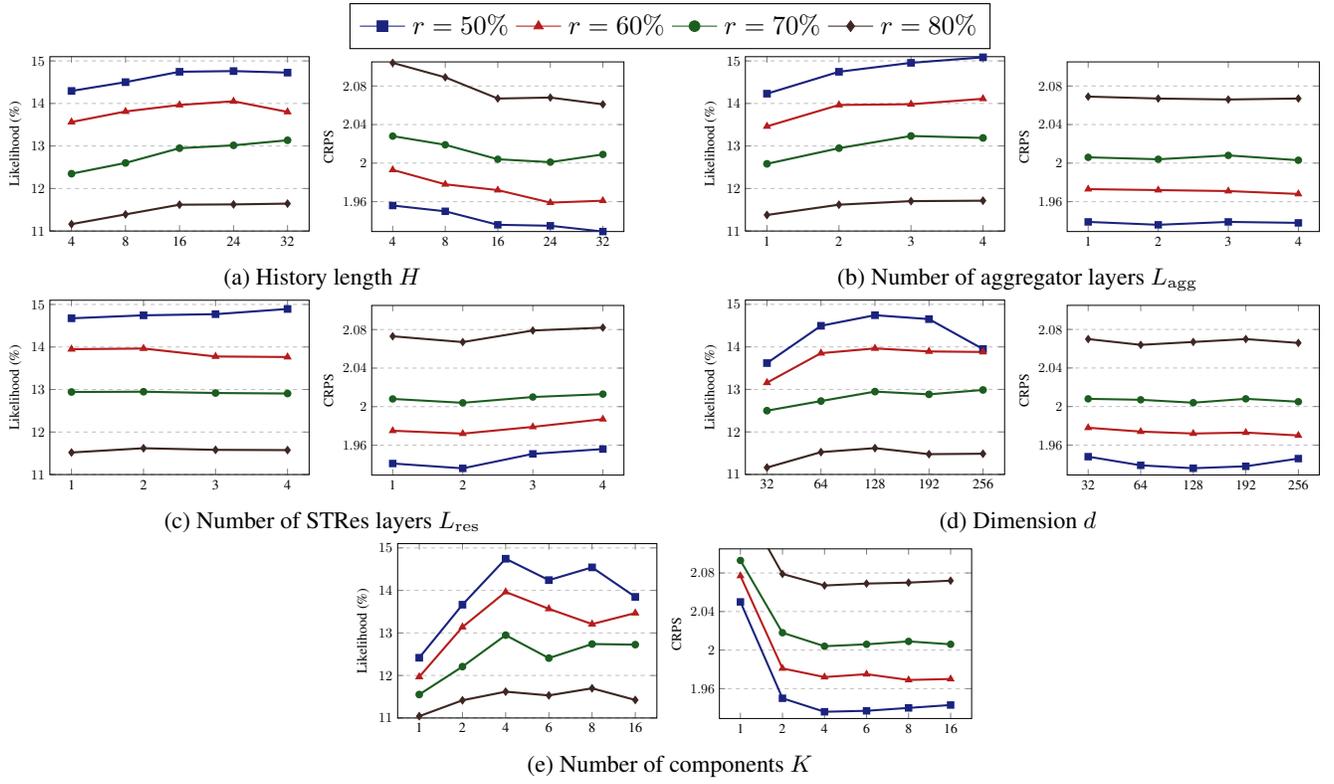
\begin{figure*}[t!]
    \centering
    \pgfplotstableread[row sep=\\,col sep=&]{
v & 50L & 60L & 70L & 80L & 50C & 60C & 70C & 80C \\
4 & 14.294 & 13.563 & 12.348 & 11.162 & 1.956 & 1.993 & 2.028 & 2.104 \\
8 & 14.501 & 13.811 & 12.601 & 11.391 & 1.950 & 1.978 & 2.019 & 2.089 \\
16 &14.745 & 13.964 & 12.948 & 11.619 & 1.936 & 1.972 & 2.004 & 2.067 \\
24 &14.761 & 14.051 & 13.015 & 11.626 & 1.935 & 1.959 & 2.001 & 2.068 \\
32 &14.725 & 13.801 & 13.134 & 11.644 & 1.929 & 1.961 & 2.009 & 2.061 \\
}\HistoryLength

\pgfplotstableread[row sep=\\,col sep=&]{
v & 50L & 60L & 70L & 80L & 50C & 60C & 70C & 80C \\
1 & 14.230 & 13.461 & 12.580 & 11.377 & 1.939 & 1.973 & 2.006 & 2.069 \\
2 & 14.745 & 13.964 & 12.948 & 11.619 & 1.936 & 1.972 & 2.004 & 2.067 \\
3 & 14.955 & 13.982 & 13.235 & 11.703 & 1.939 & 1.971 & 2.008 & 2.066 \\
4 & 15.085 & 14.110 & 13.189 & 11.713 & 1.938 & 1.968 & 2.003 & 2.067 \\
}\AggLayers

\pgfplotstableread[row sep=\\,col sep=&]{
v & 50L & 60L & 70L & 80L & 50C & 60C & 70C & 80C \\
1 & 14.675 & 13.948 & 12.942 & 11.519 & 1.941 & 1.975 & 2.008 & 2.073 \\
2 & 14.745 & 13.964 & 12.948 & 11.619 & 1.936 & 1.972 & 2.004 & 2.067 \\
3 & 14.772 & 13.777 & 12.918 & 11.581 & 1.951 & 1.979 & 2.010 & 2.079 \\
4 & 14.894 & 13.764 & 12.906 & 11.575 & 1.956 & 1.987 & 2.013 & 2.082 \\
}\ResLayers

\pgfplotstableread[row sep=\\,col sep=&]{
v & 50L & 60L & 70L & 80L & 50C & 60C & 70C & 80C \\
32  & 13.619 & 13.158 & 12.500 & 11.162 & 1.948 & 1.978 & 2.008 & 2.070 \\
64  & 14.496 & 13.852 & 12.726 & 11.525 & 1.939 & 1.974 & 2.007 & 2.064 \\
128 & 14.745 & 13.964 & 12.948 & 11.619 & 1.936 & 1.972 & 2.004 & 2.067 \\
192 & 14.653 & 13.893 & 12.884 & 11.477 & 1.938 & 1.973 & 2.008 & 2.070 \\
256 & 13.949 & 13.880 & 12.987 & 11.487 & 1.946 & 1.970 & 2.005 & 2.066 \\
}\DModel

\pgfplotstableread[row sep=\\,col sep=&]{
v & 50L & 60L & 70L & 80L & 50C & 60C & 70C & 80C \\
1 & 12.417 & 11.968 & 11.552 & 11.042 & 2.050 & 2.077 & 2.093 & 2.157 \\
2 & 13.664 & 13.139 & 12.208 & 11.416 & 1.950 & 1.981 & 2.018 & 2.079 \\
4 & 14.745 & 13.964 & 12.948 & 11.619 & 1.936 & 1.972 & 2.004 & 2.067 \\
6 & 14.242 & 13.567 & 12.408 & 11.531 & 1.937 & 1.975 & 2.006 & 2.069 \\
8 & 14.542 & 13.208 & 12.739 & 11.698 & 1.940 & 1.969 & 2.009 & 2.070 \\
16 &13.848 & 13.467 & 12.725 & 11.424 & 1.943 & 1.970 & 2.006 & 2.072 \\
}\NComponents

\newcommand{\LMin}{11}
\newcommand{\LMax}{15.1}
\newcommand{\LTick}{1}
\newcommand{\CMin}{1.929}
\newcommand{\CMax}{2.105}
\newcommand{\CTick}{0.04}
\newcommand{\plotWidth}{0.95\linewidth}
\newcommand{\plotHeight}{0.7\linewidth}

\tikzset{every plot/.style={line width=1.5pt}}

\ref{fig:shared-legend}\\
\begin{subfigure}{0.48\linewidth}
\resizebox{0.48\linewidth}{!}{
\begin{tikzpicture}
\begin{axis}[
    width=\plotWidth, height=\plotHeight,
    ylabel={Likelihood (\%)},
    ymin=\LMin, ymax=\LMax,
    ytick distance={\LTick},
    ymajorgrids=true,
    grid style=dashed,
    xtick=data,
    symbolic x coords={4,8,16,24,32},
    legend entries={$r=50\%$,$r=60\%$,$r=70\%$,$r=80\%$},
    legend to name=fig:shared-legend,
    legend columns=-1,
]
\addplot[color=r50,mark=square*] table[x=v,y=50L]{\HistoryLength};
\addplot[color=r60,mark=triangle*] table[x=v,y=60L]{\HistoryLength};
\addplot[color=r70,mark=otimes*] table[x=v,y=70L]{\HistoryLength};
\addplot[color=r80,mark=diamond*] table[x=v,y=80L]{\HistoryLength};
\end{axis}
\end{tikzpicture}}
\resizebox{0.48\linewidth}{!}{
\begin{tikzpicture}
\begin{axis}[
    width=\plotWidth, height=\plotHeight,
    ylabel={CRPS},
    ymin=\CMin, ymax=\CMax,
    ytick distance={\CTick},
    ymajorgrids=true,
    grid style=dashed,
    xtick=data,
    symbolic x coords={4,8,16,24,32}
]
\addplot[color=r50,mark=square*] table[x=v,y=50C]{\HistoryLength};
\addplot[color=r60,mark=triangle*] table[x=v,y=60C]{\HistoryLength};
\addplot[color=r70,mark=otimes*] table[x=v,y=70C]{\HistoryLength};
\addplot[color=r80,mark=diamond*] table[x=v,y=80C]{\HistoryLength};
\end{axis}
\end{tikzpicture}}
\caption{History length $H$}
\label{fig:param-H}
\end{subfigure}\hfill%
\begin{subfigure}{0.48\linewidth}
\resizebox{0.48\linewidth}{!}{
\begin{tikzpicture}
\begin{axis}[
    width=\plotWidth, height=\plotHeight,
    ylabel={Likelihood (\%)},
    ymin=\LMin, ymax=\LMax,
    ytick distance={\LTick},
    ymajorgrids=true,
    grid style=dashed,
    xtick=data,
    symbolic x coords={1,2,3,4},
]
\addplot[color=r50,mark=square*] table[x=v,y=50L]{\AggLayers};
\addplot[color=r60,mark=triangle*] table[x=v,y=60L]{\AggLayers};
\addplot[color=r70,mark=otimes*] table[x=v,y=70L]{\AggLayers};
\addplot[color=r80,mark=diamond*] table[x=v,y=80L]{\AggLayers};
\end{axis}
\end{tikzpicture}}
\resizebox{0.48\linewidth}{!}{
\begin{tikzpicture}
\begin{axis}[
    width=\plotWidth, height=\plotHeight,
    ylabel={CRPS},
    ymin=\CMin, ymax=\CMax,
    ytick distance={\CTick},
    ymajorgrids=true,
    grid style=dashed,
    xtick=data,
    symbolic x coords={1,2,3,4},
]
\addplot[color=r50,mark=square*] table[x=v,y=50C]{\AggLayers};
\addplot[color=r60,mark=triangle*] table[x=v,y=60C]{\AggLayers};
\addplot[color=r70,mark=otimes*] table[x=v,y=70C]{\AggLayers};
\addplot[color=r80,mark=diamond*] table[x=v,y=80C]{\AggLayers};
\end{axis}
\end{tikzpicture}}
\caption{Number of aggregator layers $L_\mathrm{agg}$}
\label{fig:param-Lagg}
\end{subfigure}\\
\begin{subfigure}{0.48\linewidth}
\resizebox{0.48\linewidth}{!}{
\begin{tikzpicture}
\begin{axis}[
    width=\plotWidth, height=\plotHeight,
    ylabel={Likelihood (\%)},
    ymin=\LMin, ymax=\LMax,
    ytick distance={\LTick},
    ymajorgrids=true,
    grid style=dashed,
    xtick=data,
    symbolic x coords={1,2,3,4},
]
\addplot[color=r50,mark=square*] table[x=v,y=50L]{\ResLayers};
\addplot[color=r60,mark=triangle*] table[x=v,y=60L]{\ResLayers};
\addplot[color=r70,mark=otimes*] table[x=v,y=70L]{\ResLayers};
\addplot[color=r80,mark=diamond*] table[x=v,y=80L]{\ResLayers};
\end{axis}
\end{tikzpicture}}
\resizebox{0.48\linewidth}{!}{
\begin{tikzpicture}
\begin{axis}[
    width=\plotWidth, height=\plotHeight,
    ylabel={CRPS},
    ymin=\CMin, ymax=\CMax,
    ytick distance={\CTick},
    ymajorgrids=true,
    grid style=dashed,
    xtick=data,
    symbolic x coords={1,2,3,4},
]
\addplot[color=r50,mark=square*] table[x=v,y=50C]{\ResLayers};
\addplot[color=r60,mark=triangle*] table[x=v,y=60C]{\ResLayers};
\addplot[color=r70,mark=otimes*] table[x=v,y=70C]{\ResLayers};
\addplot[color=r80,mark=diamond*] table[x=v,y=80C]{\ResLayers};
\end{axis}
\end{tikzpicture}}
\caption{Number of STRes layers $L_\mathrm{res}$}
\label{fig:param-Lres}
\end{subfigure}\hfill%
\begin{subfigure}{0.48\linewidth}
\resizebox{0.48\linewidth}{!}{
\begin{tikzpicture}
\begin{axis}[
    width=\plotWidth, height=\plotHeight,
    ylabel={Likelihood (\%)},
    ymin=\LMin, ymax=\LMax,
    ytick distance={\LTick},
    ymajorgrids=true,
    grid style=dashed,
    xtick=data,
    symbolic x coords={32,64,128,192,256},
]
\addplot[color=r50,mark=square*] table[x=v,y=50L]{\DModel};
\addplot[color=r60,mark=triangle*] table[x=v,y=60L]{\DModel};
\addplot[color=r70,mark=otimes*] table[x=v,y=70L]{\DModel};
\addplot[color=r80,mark=diamond*] table[x=v,y=80L]{\DModel};
\end{axis}
\end{tikzpicture}}
\resizebox{0.48\linewidth}{!}{
\begin{tikzpicture}
\begin{axis}[
    width=\plotWidth, height=\plotHeight,
    ylabel={CRPS},
    ymin=\CMin, ymax=\CMax,
    ytick distance={\CTick},
    ymajorgrids=true,
    grid style=dashed,
    xtick=data,
    symbolic x coords={32,64,128,192,256},
]
\addplot[color=r50,mark=square*] table[x=v,y=50C]{\DModel};
\addplot[color=r60,mark=triangle*] table[x=v,y=60C]{\DModel};
\addplot[color=r70,mark=otimes*] table[x=v,y=70C]{\DModel};
\addplot[color=r80,mark=diamond*] table[x=v,y=80C]{\DModel};
\end{axis}
\end{tikzpicture}}
\caption{Dimension $d$}
\label{fig:param-d}
\end{subfigure}\\
\begin{minipage}{\linewidth}\centering
\begin{subfigure}{0.48\linewidth}
\resizebox{0.48\linewidth}{!}{
\begin{tikzpicture}
\begin{axis}[
    width=\plotWidth, height=\plotHeight,
    ylabel={Likelihood (\%)},
    ymin=\LMin, ymax=\LMax,
    ytick distance={\LTick},
    ymajorgrids=true,
    grid style=dashed,
    xtick=data,
    symbolic x coords={1,2,4,6,8,16},
]
\addplot[color=r50,mark=square*] table[x=v,y=50L]{\NComponents};
\addplot[color=r60,mark=triangle*] table[x=v,y=60L]{\NComponents};
\addplot[color=r70,mark=otimes*] table[x=v,y=70L]{\NComponents};
\addplot[color=r80,mark=diamond*] table[x=v,y=80L]{\NComponents};
\end{axis}
\end{tikzpicture}}
\resizebox{0.48\linewidth}{!}{
\begin{tikzpicture}
\begin{axis}[
    width=\plotWidth, height=\plotHeight,
    ylabel={CRPS},
    ymin=\CMin, ymax=\CMax,
    ytick distance={\CTick},
    ymajorgrids=true,
    grid style=dashed,
    xtick=data,
    symbolic x coords={1,2,4,6,8,16},
]
\addplot[color=r50,mark=square*] table[x=v,y=50C]{\NComponents};
\addplot[color=r60,mark=triangle*] table[x=v,y=60C]{\NComponents};
\addplot[color=r70,mark=otimes*] table[x=v,y=70C]{\NComponents};
\addplot[color=r80,mark=diamond*] table[x=v,y=80C]{\NComponents};
\end{axis}
\end{tikzpicture}}
\caption{Number of components $K$}
\label{fig:param-K}
\end{subfigure}
\end{minipage}
    \caption{Effectiveness of hyper-parameters.}
    \label{fig:hyper-parameters}
\end{figure*}

We explore the impact of the five key hyper-parameters on the performance of DiSGMM. The selection range and optimal values of these hyper-parameters are detailed in Table~\ref{table:hyper-parameters}.

\begin{table}[h]
\centering
\caption{Hyper-parameter range and optimal values.}
\label{table:hyper-parameters}
\begin{threeparttable}
\begin{tabular}{c|c}
\toprule
Parameter & Range \\
\midrule
$H$ & 4, 8, \underline{16}, 24, 32 \\
$L_\mathrm{agg}$ & 1, \underline{2}, 3, 4 \\
$L_\mathrm{res}$ & 1, \underline{2}, 3, 4 \\
$d$ & 32, 64, \underline{128}, 192, 256 \\
$K$ & 1, 2, \underline{4}, 6, 8, 16\\
\bottomrule
\end{tabular}
\begin{tablenotes}\footnotesize
    \item[]{\underline{Underline} denotes the optimal value.}
\end{tablenotes}
\end{threeparttable}

\end{table}

The results, obtained on the HTT dataset, are presented in Figure~\ref{fig:hyper-parameters}.
We have the following observations:
\begin{enumerate}[leftmargin=*]
\item Expanding the history length $H$, as shown in Figure~\ref{fig:param-H}, leads to improved completion performance. This is because the method can capture temporal correlations from more distant periods. However, it also increases the computational costs. To strike a balance between performance and efficiency, we select $H=16$.
\item Increasing the number of aggregator layers, as shown in Figure~\ref{fig:param-Lagg}, enhances the completion accuracy. However, it also incurs higher memory and computational overheads. To achieve a trade-off between performance and accuracy, we choose an appropriate value for $L_\mathrm{agg}$.
\item Figure~\ref{fig:param-Lres} demonstrates that the model reaches its optimal performance when the number of STRes layers $L_\mathrm{res}=2$. Fewer layers decrease the learning capability of the model, while more layers lead to overfitting.
\item The dimension $d$ exhibits an optimal value of $d=128$, as indicated in Figure~\ref{fig:param-d}. Similar to $L_\mathrm{res}$, a smaller $d$ negatively affects the learning capability, while a larger $d$ results in overfitting.
\item Figure~\ref{fig:param-K} reveals that the optimal number of components is achieved when $K=4$. A smaller $K$ fails to adequately represent the distributions of microscopic weights, while a larger $K$ introduces noise into the estimated GMMs.
\end{enumerate}

\section{Case Study}
\label{sec:case-study}

\begin{figure*}[t]
    \centering
    \includegraphics[width=1.0\linewidth]{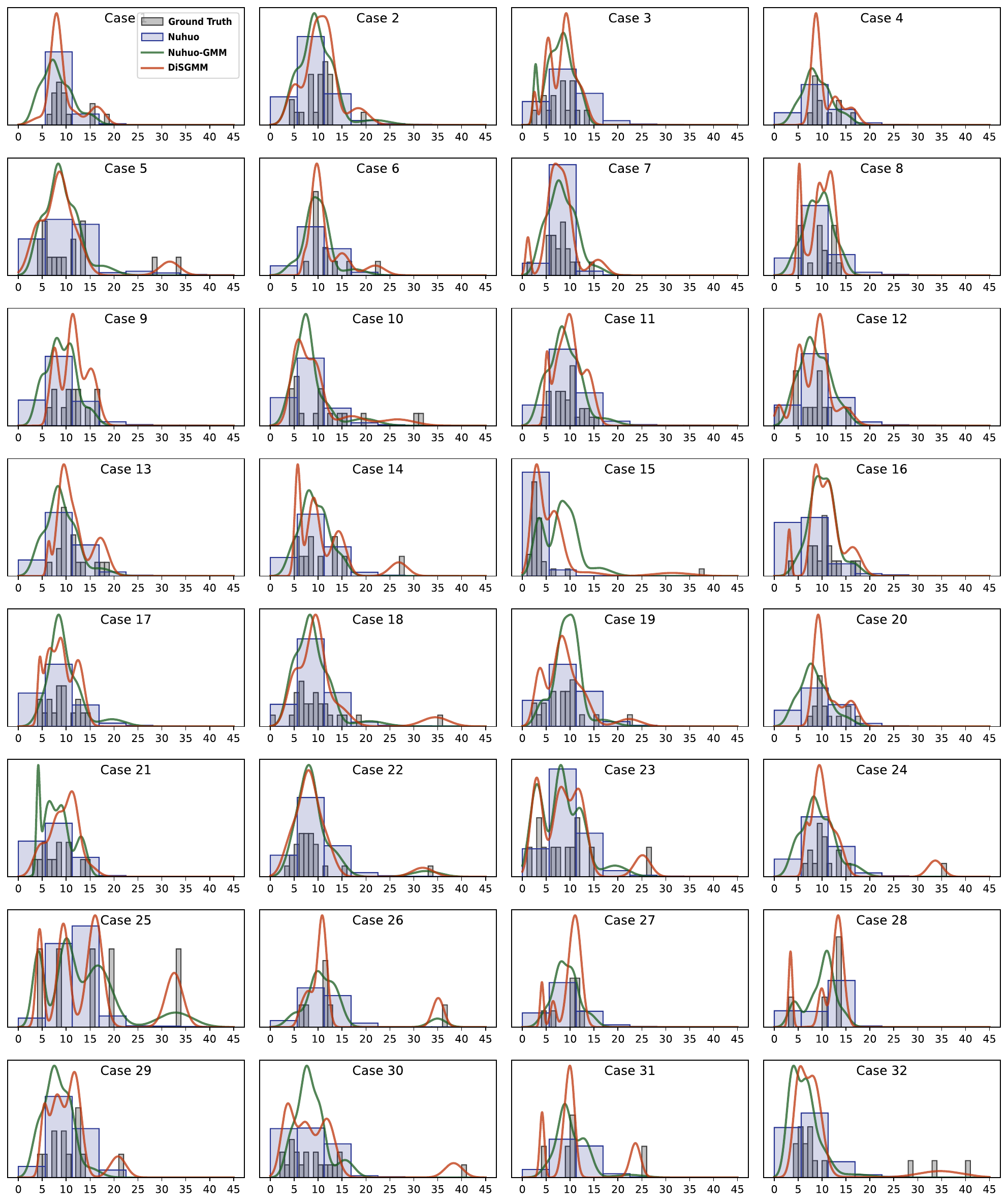}
    \caption{Visualization of known microscopic weights and estimated weight distributions from different methods.}
    \label{fig:case-study}
\end{figure*}

To provide an intuitive comparison of the estimated weight distributions, we visualize the results of Nuhuo, Nuhuo-GMM, and DiSGMM on selected road segments. Specifically, under the HTT dataset with a missing rate of $r=80\%$, we select a time period from the testing set and identify the road segment with the most known weights. The known samples are represented as a fine-grained histogram serving as the ground truth distribution. The results are presented in Figure~\ref{fig:case-study}.

We have the following observations.
First, GMMs represent the underlying distributions of microscopic weights more accurately than histograms. This demonstrates the trade-off between range coverage and resolution: histograms that cover the full range (extending to values above 30 m/s in several cases) sacrifice resolution where most values concentrate. In contrast, GMMs can adapt their means and variances to capture both dense regions with high resolution and heavy tails. Cases 1, 3, 9, and 17 in Figure~\ref{fig:case-study} exemplify this: the histogram (Nuhuo) shows coarse bins that blur nearby clusters, while the GMM curves (Nuhuo-GMM, DiSGMM) distinguish them clearly.
Second, comparing the estimated distributions of Nuhuo-GMM and DiSGMM, we find that DiSGMM produces more accurate estimations in most cases. The distributions estimated by DiSGMM align more closely with the ground truth, demonstrating its effectiveness in microscopic weight completion.

\section{Additional Implementation Details}

\subsection{Static Road Segment Embedding Details}
\label{sec:static-embedding-details}

This appendix provides the detailed formulation of the static road segment embedding module described in Section~\ref{sec:embedding-module}.

Given the road network $\mathcal G=(\mathcal V, \mathcal E)$ and a target road segment $e\in \mathcal E$, we perform random walks on $\mathcal G$ to generate sequences centered around $e$ containing $C$ contextual segments:
\begin{equation}
\mathcal S_e = \langle e_1, e_2, \dots, e, \dots, e_{C} \rangle,
\end{equation}
where $\{e_1,e_2,\dots,e_C\}$ are the contextual segments that appear together with $e$ in $\mathcal S_e$. We set $C=4$ in this study.

We also gather the historical microscopic weights of $e$:
\begin{equation}
\{\mathbb W_e^{T_j} | T_j\in \mathcal T_\mathrm{his}\},
\end{equation}
where $\mathcal T_\mathrm{his}=\{T_1,T_2,\dots,T_{N_\mathrm{his}}\}$ is the set of all historical time slots.

Drawing inspiration from skip-gram~\cite{DBLP:journals/corr/abs-1301-3781}, we maximize the appearance probability of contextual segments given the target segment $e$ and one of its historical microscopic weights $w_e$:
\begin{equation}
\mathrm{max}~p(e_1,e_2,\dots,e_C|e,w_e).
\end{equation}

We assign an embedding vector $\boldsymbol z_e\in \mathbb R^d$ to $e$ and an output vector $\boldsymbol z'_{e_k} \in \mathbb R^{2d}$ to each road segment $e_k \in \mathcal E$. We calculate a hidden vector $\Phi(w_e) \in \mathbb R^d$ for each microscopic weight using learnable Fourier features~\cite{DBLP:conf/nips/LiSLHB21}. The loss function is:
\begin{equation}
\mathcal L_{\mathrm{embed}} = -\log \prod_{c=1}^C \frac{\exp((\boldsymbol z_e || \Phi(w_e)) {\boldsymbol z'_{e_c}}^\top)}{\sum_{e_k\in \mathcal E} \exp((\boldsymbol z_e || \Phi(w_e))  {\boldsymbol z'_{e_k}}^\top)},
\end{equation}
where $||$ denotes vector concatenation.

\subsection{Sparsity-aware Dynamic Embedding Details}
\label{sec:dynamic-embedding-details}

This appendix provides the detailed formulation of the sparsity-aware dynamic embedding module described in Section~\ref{sec:dynamic-embedding}.

Given the known microscopic weights $\mathbb W_e^{T_j}$ of a target segment $e$ during time slot $T_j$, we first map each weight $w_e\in \mathbb W_e^{T_j}$ to a hidden vector using the learnable Fourier feature $\Phi$:
\begin{equation}
\begin{split}
\Phi(\mathbb W_e^{T_j}) &= \Phi(\{w_{e,1}^{T_j},w_{e,2}^{T_j},\dots,w_{e,N}^{T_j}\}) \\
&= \{\Phi(w_{e,1}^{T_j}),\Phi(w_{e,2}^{T_j}),\dots,\Phi(w_{e,N}^{T_j})\},
\end{split}
\end{equation}
where $N=|\mathbb W_e^{T_j}|$. The mapped vectors are aggregated using a Transformer encoder followed by mean pooling:
\begin{equation}
\mathrm{Agg}(\mathbb W_e^{T_j})=\mathrm{Mean}(\mathrm{TransEnc}(\Phi(\mathbb W_e^{T_j}))),
\label{eq:aggregate-weights}
\end{equation}
where $\mathrm{TransEnc}$ consists of $L_\mathrm{agg}$ Transformer encoder layers without positional encoding. For empty sets, $\mathrm{Agg}(\mathbb W_e^{T_j}) = \boldsymbol 0\in \mathbb R^d$.

The complete gate mechanism is:
\begin{equation}
\begin{split}
\boldsymbol f_e^{T_j} &= \sigma({\Psi_e^{T_j}}^\top \boldsymbol W_f \mathrm{Agg}_e^{T_j} + {\Psi_e^{T_j}}^\top \boldsymbol U_f \boldsymbol z_e + \boldsymbol b_f), \\
\hat{\boldsymbol{h}}_e^{T_j} &= \mathrm{tanh}(\boldsymbol W_h \mathrm{Agg}_e^{T_j} + \boldsymbol U_h(\boldsymbol f_e^{T_j}\odot \boldsymbol z_e) + \boldsymbol b_h), \\
\boldsymbol{h}_e^{T_j} &= (1-\boldsymbol f_e^{T_j})\odot \boldsymbol z_e + \boldsymbol f_e^{T_j} \odot \hat{\boldsymbol{h}}_e^{T_j},
\end{split}
\end{equation}
where $\Psi_e^{T_j} = \Psi(|\mathbb W_e^{T_j}|)$ encodes the sparsity level, $\boldsymbol f_e^{T_j}$ is the forget gate, $\hat{\boldsymbol{h}}_e^{T_j}$ is the candidate hidden vector, and $\boldsymbol W_f, \boldsymbol U_f, \boldsymbol b_f, \boldsymbol W_h, \boldsymbol U_h, \boldsymbol b_h$ are learnable parameters.

\subsection{Spatiotemporal U-net Implementation Details}
\label{sec:unet-details}

This appendix provides detailed formulations of the STRes and down/up-sampling blocks described in Section~\ref{sec:spatiotemporal-unet}.

\subsubsection{STRes Block}

Given a set $\mathcal T=\{T_1, T_2, \dots, T_H\}$ of time slots in temporal ascending order, the input to the first STRes block is formed by gathering the dynamic embeddings:
\begin{equation}
\begin{split}
\boldsymbol H_e &= \langle \boldsymbol h_e^{T_1}, \boldsymbol h_e^{T_2}, \dots, \boldsymbol h_e^{T_H} \rangle \in \mathbb R^{H\times d_\mathrm{in}}, \\
\boldsymbol H &= \langle \boldsymbol H_{e_1}, \boldsymbol H_{e_2}, \dots, \boldsymbol H_{e_{|\mathcal E|}} \rangle \in \mathbb R^{|\mathcal E| \times H \times d_\mathrm{in}},
\end{split}
\end{equation}
where $d_\mathrm{in}=d$ in the first STRes block.

Temporal convolution is performed with a two-layer TCN followed by layer normalization:
\begin{equation}
\boldsymbol H' = \mathrm{LayerNorm}(\mathrm{TCN}(\mathrm{TCN}(\boldsymbol H))),
\label{eq:temporal-convolution}
\end{equation}
where $\mathrm{TCN}$ is implemented with 1D convolution of kernel size 3 and padding 3. The output $\boldsymbol H' \in \mathbb R^{|\mathcal E| \times H \times d_\mathrm{out}}$.

Spatial convolution captures neighboring spatial correlations:
\begin{equation}
\begin{split}
\tilde{\boldsymbol A} &= \boldsymbol D^{-1/2}(\boldsymbol A+\boldsymbol I) \boldsymbol D^{-1/2}, \\
\boldsymbol H'' &= \mathrm{ReLU}(\tilde{\boldsymbol A} \boldsymbol H' \boldsymbol W_\mathrm{fwd} + \tilde{\boldsymbol A^\top} \boldsymbol H' \boldsymbol W_\mathrm{rev} + \boldsymbol b_\mathrm{spa}),
\end{split}
\label{eq:spatial-convolution}
\end{equation}
where $\tilde{\boldsymbol A}$ is the normalized Laplacian matrix, $\boldsymbol W_\mathrm{fwd}, \boldsymbol W_\mathrm{rev} \in \mathbb R^{d_\mathrm{out}\times d_\mathrm{out}}$ are learnable matrices for forward and reverse directions, and $\boldsymbol b_\mathrm{spa} \in \mathbb R^{d_\mathrm{out}}$ is the learnable bias.

The STRes block output includes a residual connection:
\begin{equation}
\boldsymbol {\bar H} = \boldsymbol H'' + \boldsymbol H\boldsymbol W_\mathrm{res},
\end{equation}
where $\boldsymbol W_\mathrm{res} \in \mathbb R^{d_\mathrm{in}\times d_\mathrm{out}}$ is a learnable matrix.

\subsubsection{Down/Up-Sampling Blocks}

For a given input tensor $\boldsymbol H_\mathrm{in}\in \mathbb R^{|\mathcal E| \times N_\mathrm{in} \times d_\mathrm{in}}$:
\begin{itemize}[leftmargin=*]
\item \textbf{Down-sampling}: 1D convolution with kernel size 3, padding 1, stride 2, producing output shape $\mathbb R^{|\mathcal E|\times N_\mathrm{in}/2 \times d_\mathrm{in}}$.
\item \textbf{Up-sampling}: 1D transpose convolution with kernel size 4, padding 2, stride 1, producing output shape $\mathbb R^{|\mathcal E|\times 2N_\mathrm{in}\times d_\mathrm{in}}$.
\end{itemize}

The STRes blocks in the left component have $d_\mathrm{out}=2d_\mathrm{in}$, middle component have $d_\mathrm{out}=d_\mathrm{in}$, and right component have $d_\mathrm{out}=d_\mathrm{in}/2$.

\end{document}